\documentclass[a4paper,fleqn]{cas-sc}

\usepackage[authoryear,longnamesfirst]{natbib}

\usepackage{graphicx}
\graphicspath{{Figs/}}
\usepackage{subcaption}
\usepackage{color}
\usepackage{multirow}
\newcommand{\pa}{\textcolor{black}}
\newcommand{\rv}{\textcolor{black}}
\newcommand{\rvv}{\textcolor{black}}

\def\tsc#1{\csdef{#1}{\textsc{\lowercase{#1}}\xspace}}
\tsc{WGM}
\tsc{QE}

\begin{document}
\let\WriteBookmarks\relax
\def\floatpagepagefraction{1}
\def\textpagefraction{.001}

\shorttitle{Fuel Consumption Prediction for a Passenger Ferry using Machine Learning and In-service Data: A Comparative Study}    

\shortauthors{Pedram Agand, Allison Kennedy, Trevor Harris, Chanwoo Bae, Mo Chen, Edward J Park}

\title [mode = title]{Fuel Consumption Prediction for a Passenger Ferry using Machine Learning and In-service Data: A Comparative Study} 

\tnotemark[1]
\tnotemark[2]

\tnotetext[1]{This work was supported by the \pa{National Research Council Canada (NRC) Collaborative Science, Technology and Innovation Program}. }
\tnotetext[2]{Disclaimer: This work has been accepted for publication in Ocean Engineering journal 2023 \url{https://doi.org/10.1016/j.oceaneng.2023.115271}. © 2023. This manuscript version is made available under the CC-BY-NC-ND 4.0 license \url{https://creativecommons.org/licenses/by-nc-nd/4.0/}
 }



%

\author[1]{Pedram Agand}[orcid=0000-0001-5638-585X]

\cormark[1]


\ead{pagand@sfu.ca}



\affiliation[1]{organization={Department of Computer Science},
            addressline={Simon Fraser University}, 
            city={Burnaby},
            state={BC},
            country={Canada}}

\author[2]{Allison Kennedy}[]
\affiliation[2]{organization={Ocean, Costal, and River Engineering},
            addressline={National Research Council Canada}, 
            city={Ottawa},
            state={Ontario},
            country={Canada}}

\author[2]{Trevor Harris}[]

\author[3]{Chanwoo Bae}[]
\credit{credit 1}
\affiliation[3]{organization={Naval Architecture},
            addressline={BC Ferries}, 
            city={Victoria},
            state={BC},
            country={Canada}}

\author[1]{Mo Chen}[]

\author[4]{Edward J. Park}[]
\affiliation[4]{organization={School of Mechatronic Systems Engineering},
            addressline={Simon Fraser University}, 
            city={Surrey},
            state={BC},
            country={Canada}}

\cortext[1]{Corresponding author.}



\begin{abstract}
As the importance of eco-friendly transportation increases, providing an efficient approach for marine vessel operation is essential. Methods for status monitoring with consideration to the weather condition and forecasting with the use of  in-service data from ships requires accurate and complete models for predicting the energy efficiency of a ship. The models need to effectively process all the operational data in real-time. This paper presents models that can predict fuel consumption using in-service data collected from a passenger ship. Statistical and domain-knowledge methods were used  to select the proper input variables for the models. These methods prevent over-fitting, missing data, and multicollinearity while providing practical applicability. Prediction models that were investigated include multiple linear regression (MLR), decision tree approach (DT), an artificial neural network (ANN), and ensemble methods. The best predictive performance was from a model developed using the XGboost technique which is a boosting ensemble approach. \rvv{Our code is available on GitHub at \url{https://github.com/pagand/model_optimze_vessel/tree/OE} for future research. }
\end{abstract}
\maketitle
\section{Introduction}\label{intro}
More than 80\% of global goods commerce is transported by sea, making the shipping sector one of the foundations of the global economy \cite{sudi2021effect}. Compared to other forms of transportation, carbon dioxide emissions from shipping constitute a significant portion of the world's overall greenhouse gas emissions \cite{becsikcci2016artificial}. The International Maritime Organization (IMO) has accepted a proposed amendment making it mandatory for all ships to have a SEEMP (Ship Energy Efficiency Management Plan) and an EEDI (Energy Efficiency Design Index) for new ships. This implementation aims to reduce greenhouse gas emissions from international commercial vessels and is currently in place for ships weighing at least 400GT \cite{joung2020imo}. Fuel prices have surpassed all other operating expenses in the shipping industry, drawing the attention of shipping companies and government regulators \cite{branch2013maritime}. Ship Fuel Consumption (SFC) accounts for over 25\% of a vessel's operational costs \cite{gkerekos2019machine}, almost two-thirds of its voyage expenditures \cite{theodoridis2009chapter}, and approximately 50\% to 60\% of all operating costs \cite{eide2011future}.

Enhancing navigational performance can be achieved through direct management techniques employed by ship operators, such as setting an appropriate autopilot mode, employing optimal draught and trim settings, and planning routes with consideration for weather and sea conditions \cite{abs22, rudzki2016decision}. Quantitative assessments of SFC data play a significant role in environmental protection and energy management in maritime operations. Models that forecast a ship's energy efficiency in real-time must efficiently handle operational data and be tailored for such applications.

Predictive model development using log-based and sensor-based SFC data has been conducted using various techniques and processes. In the literature, model development using log-based data typically involves three steps: 1) Removing outliers from SFC data by considering statistical model noises, including univariate and multivariate outliers \cite{zhou2022adaptive}.
2) Deleting SFC data recorded at speeds of less than five knots or when no cargo is carried, followed by constructing models using Random Forest Regression (RFR) to maximize navigation speed while minimizing fuel usage and ensuring punctual arrival \cite{yan2020development}.
3) Making three initial modifications to the draught, weather, and hull roughness of the recorded SFC data and using polynomial regression analysis to illustrate the correlations between fuel usage and speed under various weather circumstances \cite{bialystocki2016estimation}. However, log-based data, despite its straightforward format, cannot accurately reflect the fuel consumption situation due to its lower sample frequency compared to sensor-based data, as indicated in the literature.

For sensor-based SFC, onboard sensors such as the Automated Identification System (AIS) are used to collect data. The following five steps briefly describe the model development using sensor-based data: 1) Completing data normalization to expedite convergence of the Artificial Neural Network (ANN) \cite{zheng2019voyage}.
2) Using the mean, variance, and mean difference as features in the process \cite{petersen2011mining}.
3) Enhancing navigation speed with on-time arrival using a dynamic programming (DP) technique \cite{zhu2021modeling}.
4) Considering variable redefinitions, such as converting the wind's direction into headwind and crosswind, and employing feature selection to ensure the use of proper regressors \cite{hastie2009elements}.
5) Linking fuel consumption to ship operation activities and employing the Kalman filter to clean up corrupted data collected when the ship is not in motion \cite{trodden2015fuel}. In \cite{hu2022two}, a two-step method is suggested for the forecast and optimization of ship fuel consumption. Fuel consumption data is first analyzed, and a unique hybrid prediction model based on the stacking theory is created by combining various cutting-edge single models. The second stage suggests a way to optimize fuel use from the standpoint of trimming.

\pa{Despite the numerous studies conducted, the main gap in the literature is the reliance solely} on the expertise of personnel onboard the ship, leading to discrepancies between the practical needs of navigational management and the predictions of fuel consumption models. Earlier studies predominantly focused on optimization at the vessel design level without properly considering improvements in operational performance from the ship operator's perspective. Therefore, the model's input variables must include the ship operator's real-time input elements. Additionally, environmental components should be taken into account during modeling, as external disturbances are inevitable in decision-making. \pa{Furthermore, there is a lack of study on feature selection in developing an analytical model}. Such studies are crucial because many characteristic factors related to the ship's operational performance exhibit significant correlations among themselves, while others may have a weaker effect on the target variable. The presence of multicollinearity between variables can lead to improper estimation of regression coefficients if these characteristic variables are directly employed in a prediction model.

Another challenge in using machine learning models on operational data is ensuring the data's stationarity, as different circumstances may result in data from different distributions. For instance, changes in the ship's captain and the mode of operation (docking zones and cruising/autopilot phases) can lead to non-stationary data. \pa{However, little to no research has been conducted on developing a systematic approach to select relevant parts of datasets other than manually selecting or cropping the data based on hand-selected criteria.}

An important consideration when applying machine learning to operational data is the integration of domain knowledge with data-driven approaches. While models solely based on physical equations are ineffective, domain knowledge can enhance performance. Previous works either solely used physical models as a basis or solely relied on data-driven approaches using machine learning models. The systematic combination of these perspectives was lacking in earlier research.

Another weakness in the literature is the lack of comparison of $n$-step ahead prediction performance. \rv{
In the field of prediction, the concept of n-step ahead prediction refers to forecasting the target variable at a future time point, specifically at time $t+n$, based on the available information up to the current time point, denoted as time $t$. This particular prediction task is commonly recognized as more challenging and necessitates the utilization of more intricate models.}
While most models aim to estimate instantaneous fuel consumption, the models must also perform adequately in predicting the target variable for future time points to be useful for optimization purposes. This aspect is crucial for fuel optimization.

In this paper, we demonstrate approaches to address common issues in non-stationary data, multicollinearity, outlier detection, missing/corrupted data, and data normalization during the data preprocessing stage. We employ three different criteria to address the feature selection issue. First, we incorporate domain knowledge to incorporate information from physical/operational sources into the problem, such as mapping the target variable from fuel consumption to fuel efficiency for feature engineering. Second, we compute the correlation matrix to identify collinearities between features and determine effective features on the target variable. Third, we use Principal Component Analysis (PCA) and K-means clustering to distinguish the effect of changes in operational mode. In the modeling stage, we adopt two approaches: statistical analysis and machine learning methods. Statistical analysis provides insights about the data to enrich our domain knowledge, which can be utilized in the machine learning methods. Machine learning methods encompass various parametric and non-parametric approaches applied to develop suitable prediction models. We propose a comparative study of different fuel consumption prediction models for a double-ended ferry operating on the west coast of Canada using machine learning algorithms. \rv{While the techniques are applicable to other vessels and similar prediction models in general, the details of the proposed approach and the learned model are specifically applicable to the Canada West Coast ferry.} We compare the running time, error, and future step prediction error of each model. Additionally, we compare the $n$-step ahead prediction of the models to gain better insight into the applicability of each model for optimization purposes.

The remainder of this paper is organized as follows: In Section 2, we present preliminaries for machine learning approaches, providing a list of approaches applied to SFC. Section 3 presents domain knowledge in terms of vessel specifics and physical relationships. Step-by-step data preprocessing is then presented in Section 4. Section 5 is devoted to modeling using statistical and machine learning approaches. Finally, the paper concludes in Section 6.


%

\section{Preliminaries}\label{pre}
We briefly introduce several machine learning (ML) methods in this section. We use an operational dataset, which includes different columns known as ``variables'' and different rows as entries called ``observations''. ML problems can be divided into two categories: supervised and unsupervised learning. Firstly, we present an unsupervised technique. Then, we briefly cover some well-known supervised learning methods for modeling non-dynamical systems.

Linear regression (LR), its variant (e.g., polynomial regression), and its regularized versions (e.g., Lasso and Ridge) are presented as the simplest parametric approaches. \pa{For complex models, one can use ANN as a powerful parametric approach for prediction. Non-parametric approaches such as decision trees (DT) can provide relatively better performance in prediction}. Ensemble techniques combine different approaches and may also provide better overall performance. Bagging (e.g., RFR) and boosting (e.g., Ada-Boost, gradient boosting, and XGBoost) are two ensemble approaches that will be briefly presented.

\subsection{Unsupervised learning}
In unsupervised learning, there is no labeled data available, and processing is done based on similarity in the pattern or structure of the data. In the next subsection, the clustering problem is examined as an example of an unsupervised learning approach, and we will present K-means as an approach to employ it. To have a successful ML method to predict SFC, one needs to use input variables to calculate features that are significant in terms of their effect on the SFC for the vessel of interest. We present Principal Component Analysis (PCA) as a feature selection technique to reduce the dimensionality of the data. In this paper, we use PCA to systematically cluster the operational dataset for modeling.

\subsubsection{K-means clustering }
Clustering is an unsupervised machine learning algorithm that recognizes patterns without specific labels and clusters the data according to the features. To employ clustering more effectively, one can use dimension reduction techniques before applying it. K-means is a popular method to employ clustering. Based on the number of clusters chosen (k), K-means clustering assigns a set of centroids. Each data point is assigned to the cluster with the closest centroid. The technique seeks to reduce the feature's squared Euclidean distance from the cluster centroid to which it belongs. The performance of K-means clustering is measured by the inertia, which also provides a sense of how coherent the various clusters are. Inertia is defined as follows:
\begin{equation}
I = \sum_{i=1}^N (x_i - C_k)^2,
    \label{eq:inertia}
\end{equation}
where $N$ is the number of samples within the dataset, $C$ is the center of a cluster. A model is better if its inertia is lower and has fewer clusters. This is a trade-off, though, as inertia diminishes as $k$ rises. \pa{In the ``elbow'' technique, one can determine the ideal $k$ for a dataset by finding the point where the inertia drop starts to loosen \cite{thorndike1996belong}}.

\subsubsection{Principal component analysis (PCA)}
A technique for reducing the dimensionality of a dataset while maintaining the majority of its variation is PCA. It is characterized as an orthogonal linear transformation that shifts the data into a new coordinate system such that the largest variance by some scalar projection of the data comes to lie on the first coordinate (referred to as the first principal component), the second-largest variance on the second coordinate, and so on. PCA with RFR is recommended by the author in \cite{zhu2021modeling} as a feature selection step for SFC prediction.

\subsection{Supervised learning}
For supervised learning, we have access to the labeled dataset. The two main types of modeling methodologies are parametric and non-parametric. Parametric modeling techniques are conducted using a finite set of parameters, such as LR, ANN, and SVR. Non-parametric models, in contrast, presuppose that no limited number of parameters can be used to determine the dataset distribution (e.g., DTR, RFR, and SVR with RBF). In these techniques, the amount of information that parameters can capture in the dataset increases as more training data points are added.

\subsubsection{Linear regression  (LR)}
An LR model is a parametric approach that can be defined as follows:

\begin{equation}
    \hat{y}(x,w) = w_0 + w_1 x_1 + \cdot + w_D x_D = w_0 + \sum_{j=1}^D w_j x_j
\end{equation}

where $w_i$  is a set of parameters, $\hat{y}$ is the predicted target variable, and $x_j$ are the features.  The optimal solution in the presence of Gaussian noise is obtained by minimizing the mean squared error (least squares method):
\begin{equation}
    w^* = \arg \min_w \Big\{ \sum_{i=1}^N \Big(y_i - w_0 -\sum_{j=1}^D w_j x_{ij}\Big)^2 \Big\}
    \label{eq:1}
\end{equation}

\subsubsection{Polynomial regression (PR)}
PR is an extension of the LR method and is applied to cases where the features appear in a polynomial function rather than a basic weighted linear form: 
\begin{equation}
    y(x,w) = w_0 + w_1 f_1(x_1) + \cdot + w_D f_D(x_D) = w_0 + \sum_{j=1}^D w_j f_j(x_j)
\end{equation}
Multiple linear regression, polynomial regression, ridge regression, and Lasso regression methods were extensively used as a method to predict vessel fuel consumption in previous studies (e.g., \cite{wang2018predicting, uyanik2020machine, bialystocki2016estimation}).

\subsubsection{Artificial neural network (ANN)}
ANN is a parametric approach that will find the nonlinear relation using the neurons. A type of feed-forward ANN is called a multilayer perceptron (MLP). An input layer, a hidden layer, and an output layer are the three node layers that make up an MLP. Each node, except for the input nodes, is a neuron that employs a nonlinear activation function. Backpropagation is a supervised learning method used by MLP for training \cite{agand2016transparent}. For SFC, there have been grey-box models as they combined physical knowledge with ANN \cite{leifsson2008grey}. Advanced ANN, like recurrent networks, was used for this end as well. Authors in \cite{panapakidis2020forecasting} use LSTM with an Elman neural network to forecast fuel consumption of passenger ships. A comparative study also showed ANN is better than SVR and extra tree regression (ETR) \cite{gkerekos2019machine}. The authors of \cite{zheng2019voyage} provide an SFC optimization framework. The goal function is to reduce fuel usage by taking into account station arrival time limits, unpredictability in sailing speed, and load while at sea. In order to maximize the sailing speed between stations and achieve both economic and environmental cost optimization throughout a journey, the developed ANN model is incorporated into four upgraded particle swarm optimization algorithms with global search capacity. A container ship's required shaft power or main engine fuel consumption was calculated in a research by \cite{karagiannidis2021data} under arbitrary circumstances. Two methods are described for this topic, with a focus on the statistical assessment and pre-processing of the data. Additionally, cutting-edge methods for ANN training and optimization are used. The findings show that the model's accuracy may be greatly improved with careful filtering and preparation.

\subsubsection{Regularization}
Ridge and Lasso are two prevalent regression approaches to tackle the overfitting problem \cite{agand2022human}. This phenomenon happens when the model variance is relatively high, and the model is predicting the noise instead of the underlying relation. By adding the norm of weights to the loss function, we are penalizing large values for weight, which makes the model less sensitive to noise. The solution in Eq. (\ref{eq:1}) is adjusted as follows:

\begin{equation}
    w^* = \arg\min_w \Big\{ \sum_{i=1}^N \Big(y_i - \hat{y}_i\Big)^2 + \lambda_1 \sum_{j=1}^D w_j^2 + \lambda_2 \sum_{j=1}^D \|w_j\|\Big\}
    \label{eq:2}
\end{equation}
where $y$ is the target variable, and $\lambda_1, \lambda_2 \ge 0$ are tuning parameters that control the amount of shrinkage. Ridge regression (L2 regularization) involves setting $\lambda_1 > 0$, and tries to minimize the error while maintaining the weights small; this also corresponds to assuming a zero-mean Gaussian prior on the model parameters. Lasso (L1 regularization) involves setting $\lambda_2 > 0$, and encourages the weights of non-informative features to shrink to zero; this also corresponds to assuming a zero-mean Laplacian prior on the model parameters \cite{agand2023online}.   

\subsubsection{Decision tree (DT)}
As mentioned before, DT is a type of non-parametric approach. DT partitions the feature space into rectangles and learns a simple (e.g., constant) model in each of those sections \cite{hastie2001elements}. Until the stopping rule is applied, a greedy algorithm is employed to discover the best splitting. The nodes either have criteria [decision nodes] or results [end nodes], and the branches and edges describe the node's outcome. These approaches will be used when there is little to no knowledge about the underlying dynamic or form of function that generates the output. In this case, by selecting a class of functions (a.k.a. kernels), different forms of that basis are used to predict the output. Therefore, non-parametric approaches are harder to train. In other words, they make no assumptions about a parametric form and offer a considerable deal of flexibility in the potential form of the regression curve. The assumption in DTs is that the regression curve is a collection of infinitely many different functions. It largely relies on the experimenter to provide just qualitative data about the function and let the results speak for themselves regarding the precise shape of a regression curve. In \cite{zhou2022adaptive}, an adaptive hyper-parameter tuning approach is suggested, and the impact of marine environmental conditions on fuel consumption is considered. Through testing, the suggested approach was verified using the least absolute shrinkage and selection operator (Lasso), support vector regression (SVR), RFR, and ANN. They also verified that the accuracy of fuel consumption predictions may be successfully increased by using Bayesian optimization fort hyper-parameter adjustment.

\subsubsection{Ensemble techniques}
In Ensemble methods, a model that is more effective than the original model is created by combining several algorithms or the same method in different ways. \pa{These can be applied to both parametric and non-parametric approaches; we only consider an ensemble of non-parametric methods.} To compare the best outcome, the authors in \cite{peng2020machine} employed gradient boosting regression (GBR), RFR, linear regression (LR), and k-nearest neighbor regression. The authors in \citet{hu2021novel} establish a hybrid fuel consumption prediction model that combines the methodologies of extremely randomized trees (ET), random forests (RF), and multiple linear regressions (MLR). To predict SFC, the authors in \cite{meng2016shipping} utilize the trust region algorithm. There are a few statistical methods for predicting SFC that employ maximum likelihood estimate \cite{bocchetti2015statistical}.

Bagging is a homogeneous weak learner model that combines the individual learnings for the purpose of calculating the average model. The bagging technique is extended by the RF algorithm, which uses feature randomness in addition to bagging to produce an uncorrelated forest of decision trees. The random subspace approach, also known as feature bagging, creates a random subset of features that guarantees minimal correlation between decision trees. This is the main distinction between DTs and RFs. RFs merely choose a portion of those feature splits, whereas decision trees take into account all potential feature splits. Using a technique known as bagging, RF constructs whole decision trees concurrently from random bootstrap samples of the dataset. The final prediction is calculated by averaging all of the decision tree projections.

Boosting and bagging both use a homogeneous weak learner model; however, they operate in distinct ways. In this strategy, learners acquire knowledge sequentially and adaptively to enhance learning algorithm predictions. Gradient boosting is an ensemble technique that successively adds and weights our trained predictions. However, this strategy minimizes the loss when adding the most recent prediction instead of reweighting the classifiers after each iteration by fitting the new model to fresh residuals of the prior prediction. In the end, gradient boosting refers to the process of upgrading your model via gradient descent. This decision tree boosting approach is specially implemented in XGBoost with an additional custom regularization term in the goal function. Gradient boosting is the idea of improving or ``boosting'' a single weak model by fusing it with a number of additional weak models to get a model that is collectively strong. A gradient descent algorithm over an objective function in gradient boosting is used to create the technique for additively building weak models. Gradient boosting assigns certain outcomes to the following model to minimize errors.

\section{Domain knowledge}
This section first provides an overview of the vessel information that is used as a test case throughout the paper. Next, a few general physical relations for modeling fuel consumption in the literature are reviewed from which we can \pa{supplement our domain knowledge}.

\subsection{Canada west coast ferry}
\begin{figure}[t] 
  \begin{subfigure}{.49\textwidth}
    \centering
    \includegraphics[width=.8\linewidth,trim={3cm  3cm 2.5cm 0.2cm},clip]{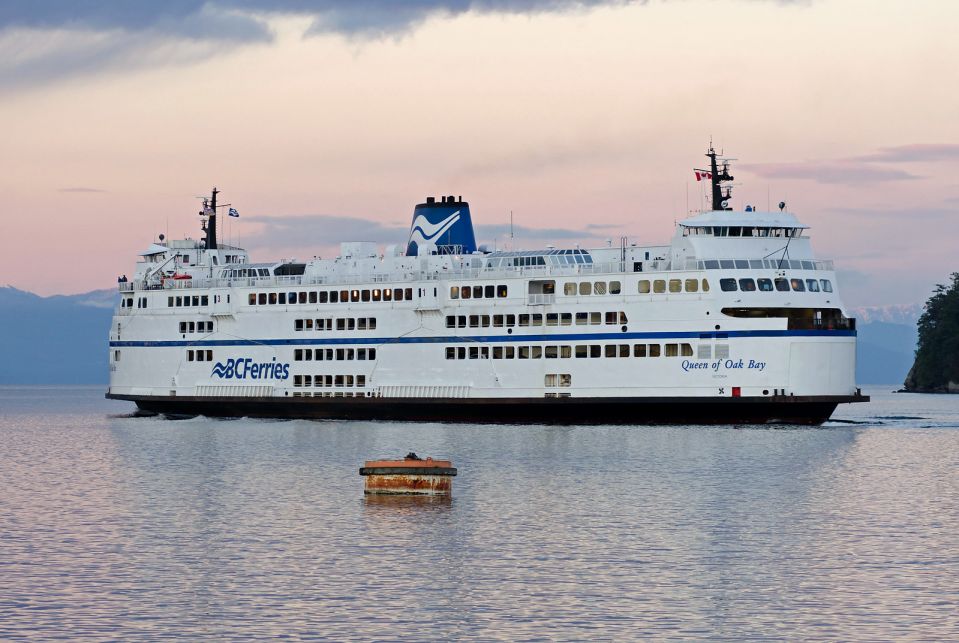} 
    \caption{Outside view \cite{bc20}}
    \vspace{1ex}
  \end{subfigure}
  \begin{subfigure}{.49\textwidth}
    \centering
    \includegraphics[width=.8\linewidth,trim={0cm  0cm 0cm 3cm},clip]{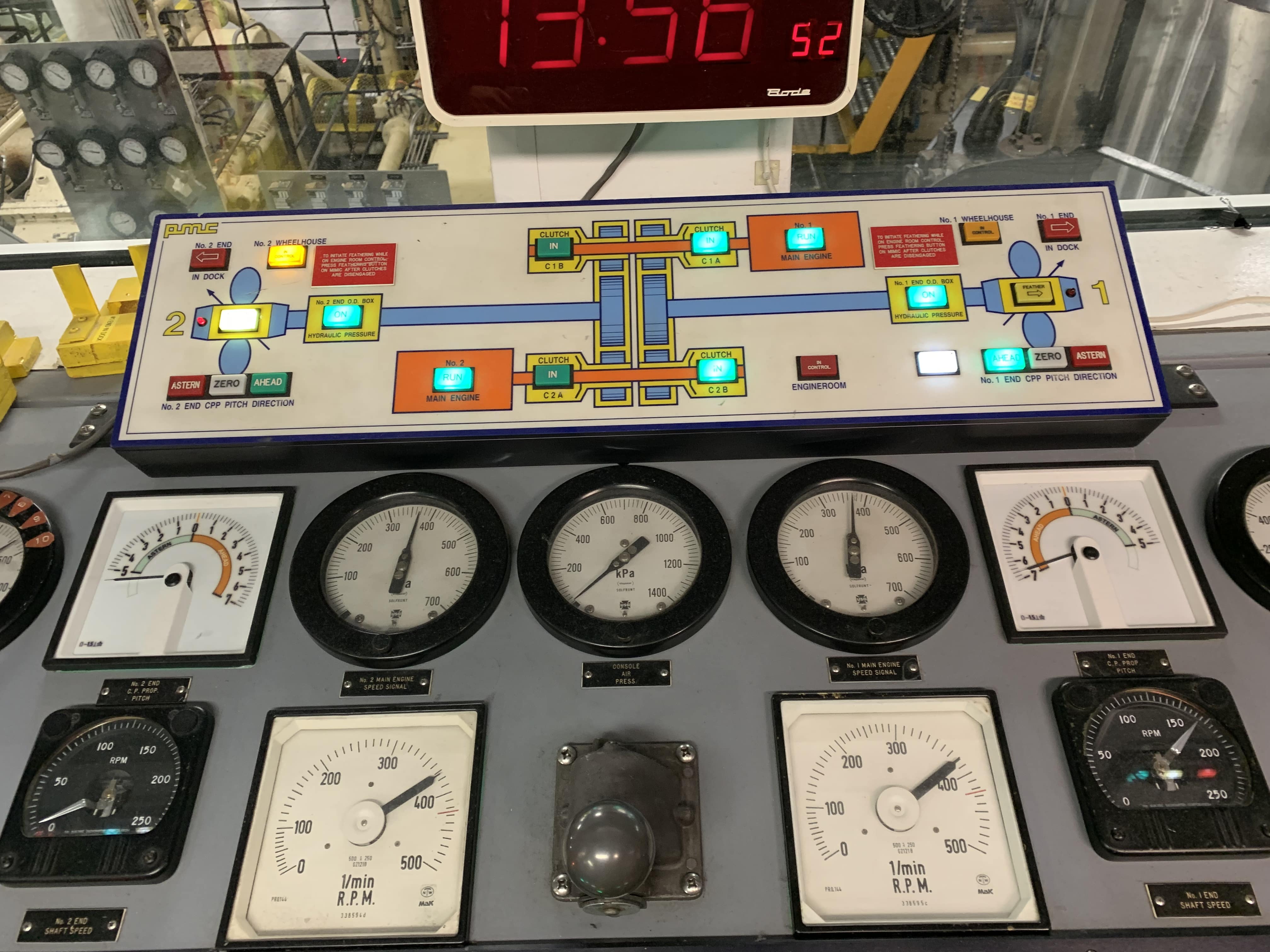}
    \caption{Engine room}
    \vspace{1ex}
 \end{subfigure}
  \begin{subfigure}{.49\textwidth}
    \centering
    \includegraphics[width=.8\linewidth,trim={0cm  0cm 0cm 2cm},clip]{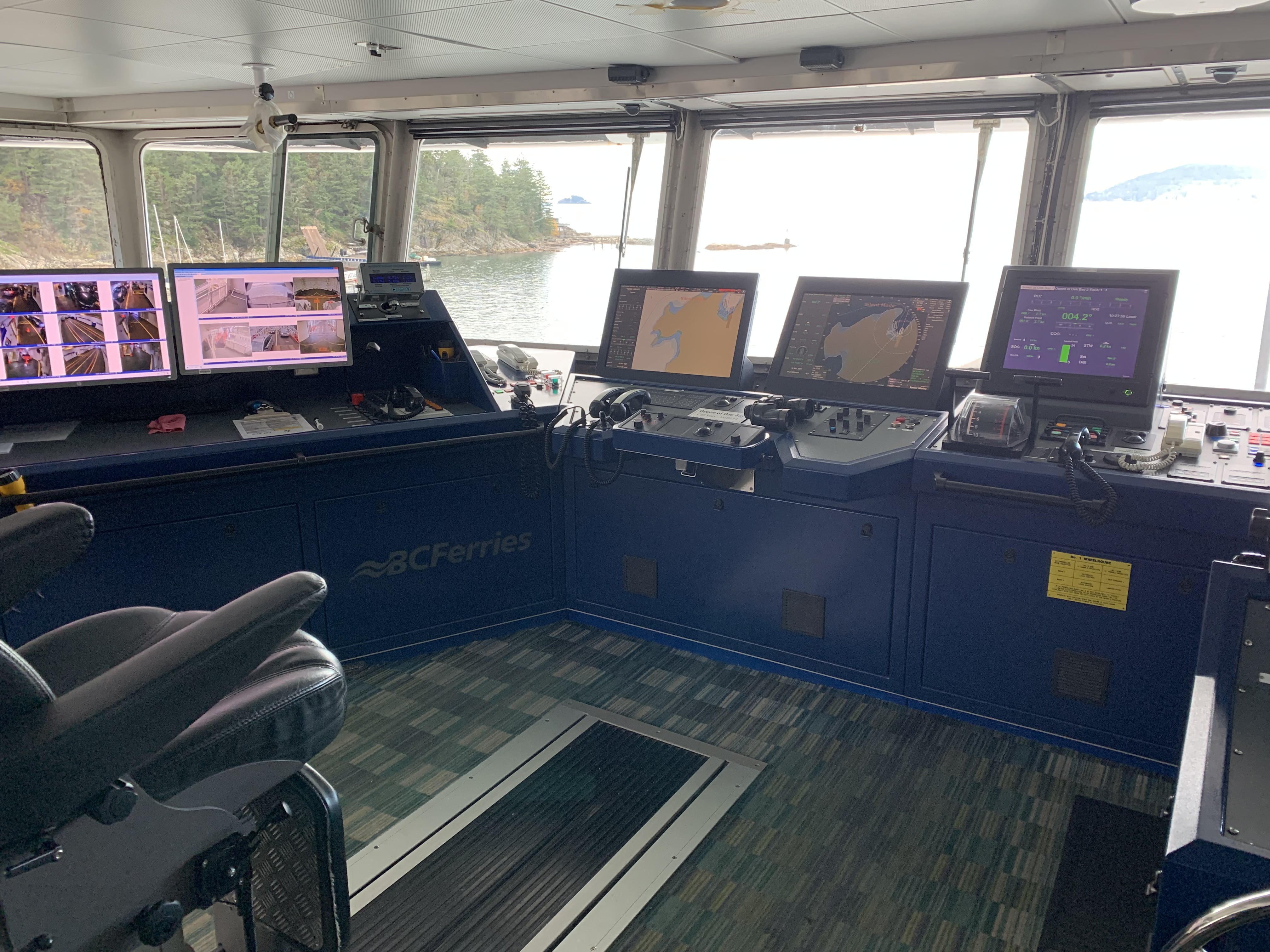} 
    \caption{Port cabin (bridge): left view}
    \vspace{1ex}
  \end{subfigure}
  \begin{subfigure}{.49\textwidth}
    \centering
    \includegraphics[width=.8\linewidth,trim={0cm  0cm 0cm 2cm},clip]{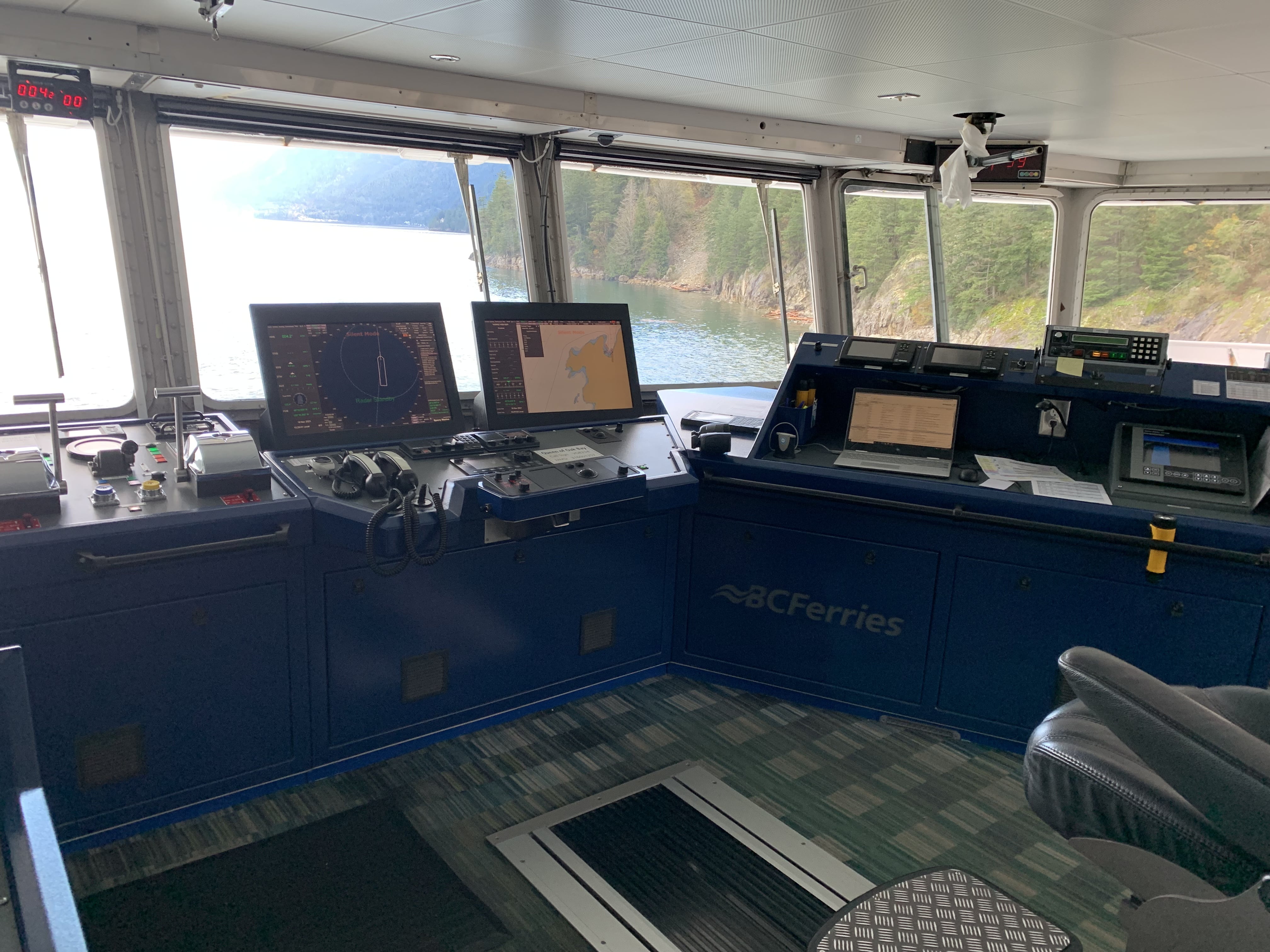} 
    \caption{Port cabin (bridge): right view}
    \vspace{1ex}
  \end{subfigure}
  \caption{The west coast Canada passenger vessel} 
  \label{fig:bcf}
\end{figure}

A picture of the actual double-end vessel is shown in Fig. \ref{fig:bcf}. The engine room is shown in Fig. \ref{fig:bcf}(b), and the bridge area is shown in Fig. \ref{fig:bcf}(c,d). This structure is identical in the bow and stern areas of the vessel. The main specifications of the vessel are illustrated in Table 1. The list of sensors available to the captain dashboard includes, but not limited to, drought control measure draft, electronic chart (GPS data, wind, depth of water), radar (360 surveillance sweep), and AIS. The actuators in the bridge include stern wheel, telegraphic control (in case of failure), jog lever (forward and aft), clutch, and speed pilot. \rv{A detailed description of the data acquisition and vessel data is presented in \cite{harris2021low}.} The duration of operational data that is available for the vessel which was used in modeling is slightly more than two years (September 2019 to November 2021). The data frequency is 1 minute (for all parameters), and each parameter was measured using a different sensor. The ferry operates on a consistent route, which involves navigating back and forth between two docking points: Horseshoe Bay and Nanaimo, British Columbia, Canada. There are a total of 41 unique variables, measured/calculated during vessel operations, and approximately 1 million data points. \rv{Using domain knowledge, we have omitted 7 columns that we knew were redundant or irrelevant, resulting in our total captured variables from 41 to 34.} The ship has two operational modes. Mode 1, or cruising phase, is when the ship is in auto-pilot, and only one of the propellers is actively engaged. Mode 2 occurs near the docking regions, during which the captain manually navigates the vessel, and both propellers are involved in different directions. There are a total of five captains that operate the ship on rotation, and the crew operates in two shifts per day.

To obtain a successful model for the vessel, a gray-box approach, which uses a combination of a physical model and mathematical methods, is suggested. The rationale behind this approach is that the whitebox model will keep the information about a ship's physical behavior, such as the equilibrium between propulsion power and vessel resistance, while the black-box model will scale the output from the white-box model to fit operational data for a specific ship and include the effects of phenomena that are not modeled in the white-box model. Therefore, the gray-box model ought to produce an accurate representation of the ship's performance that can be used for offline and real-time operational improvement \cite{nelles2020nonlinear}.

\begin{table}
\caption{Principal dimensions/specification of the vessel.}
\begin{tabular*}{\tblwidth}{@{}L@{}L|@{}L@{}L}
\toprule
 Parameter & Value&  Parameter & Value\\ 
\midrule
Length & 140 [m] & Width & 27.5 [m] \\
Draft & 27 [m] & Total Height & 32 [m] \\
Engine max power & 2$\times$8850 [kW] $~~~~~~~~$ & Propeller type & Controllable Pitch Propeller (CPP)\\
Speed of 100\% pitch &21.5 [knots] & Speed of 30\% pitch & 9 [knots] \\
\# cylinder & 12 & \# engines, propellers& 2\\
\# clutch & 4 & Shafts ratio & 2:1\\
\bottomrule
\end{tabular*}
\label{tab:bcf}
\end{table}

\subsection{Physical models for ship's fuel consumption }
According to FAO's fuel consumption calculation method, the following formula can be used to estimate liters of fuel used per machine hour \cite{dykstra1996forest}:
\begin{equation}
    LMPH = \frac{K \times GHP \times LF}{KPL}
\end{equation}
where $LMPH$ is the liters used per machine hour, $K$ is kg fuel used per brake hp/hour, $GHP$ is the gross engine horsepower at governed engine $RPM$, $LF$ is the load factor in percent, and $KPL$ is the weight of fuel in $kg/L$. The load factor is the ratio of the used average horsepower to gross horsepower available at the flywheel.

According to \citet{soleymani2018linear}, the power required to move a displacement hull through the water at velocity $V$ is proportional to $V^3$. In addition, \cite{gorski2013influence} shows that the level of fuel consumption is mostly influenced by the vessel speed as follows:
\begin{equation}
    ZP_c = \frac{\Delta^{2/3} V^3 }{ZP},
\end{equation}
where $ZP_C$, $\Delta$, $V$, and $ZP$ are fuel coefficient, ship displacement, ship speed, and main engine fuel consumption. Borkowski \cite{borkowski2011assessment} argues that ship fuel oil consumption is proportional to ship engine power. The resistance and SFC are increased with any changes in displacement (movement) \cite{bialystocki2016estimation, meng2016shipping}.

The ship's resistance is typically broken down into three main parts: 1) resistance in still water (including additional resistance owing to hull fouling), 2) additional resistance due to surface waves, and 3) resistance to wind. \rv{It is worth mentioning that the bio-fouling of a ship's hull can have a significant impact on its performance, affecting its speed, fuel efficiency, and emissions \cite{nielsen2019impact, tarelko2014effect}. This is due to a rough surface that can increase the friction between the hull and the water, which is created when marine organisms attach themselves to the hull. Studies show that bio-fouling can increase a ship's fuel consumption by up to 30\% if left untreated \cite{noufal2016impact, farkas2020impact}. Additionally, the trim of a ship can also affect its wave-making resistance, with a bow-down trim increasing resistance and a stern-down trim reducing resistance. However, the employed techniques in this paper} offer an approximation of the ship's power demand in calm and steady water and are meant to be used during conceptual design. A model that can forecast fuel consumption was provided by \cite{kim2021development}, employing in-service data as well as statistical and domain-specific methodologies to choose the right input factors for the models. ANN-based models had the greatest prediction accuracy for both variable selection approaches when MLR was used to create the prediction model. According to the sensitivity analysis of the draft under typical operating conditions, an ideal draft was extremely near to the target ship's design draft.

 
\section{Data prepossessing stage}
This section provides a summary of the data processing required to prepare the dataset for model development. This involved the implementation of different data processing techniques to remove anomalies, cluster the data, and identify significant features to include in the model. The following generic and application-specific stages were taken with respect to the operational dataset received from the vessel to provide an appropriate dataset for training.

\subsection{Clustering operational data}
As mentioned earlier, the vessel has two operational modes: Mode 1 and Mode 2. Although there is no sensory input in our dataset to distinguish these modes in the system, knowing the mode is essential since it drastically affects the fuel consumption performance of the vessel and, as such, the underlying model of the system. In addition, vessel handling by different captains and changes in the draft from the dead-weight may lead to non-stationary data. This could deteriorate the performance of the prediction model. Therefore, we employed clustering approaches with PCA to identify the group of data that are similar and then fit models only to the one cluster of interest.

\begin{figure*}[t!] 
\begin{center}
  \begin{subfigure}[t]{0.49\textwidth}
    \centering
    \includegraphics[height=2.3in,trim={1.48cm  .1cm 1.5cm 1.2cm},clip]{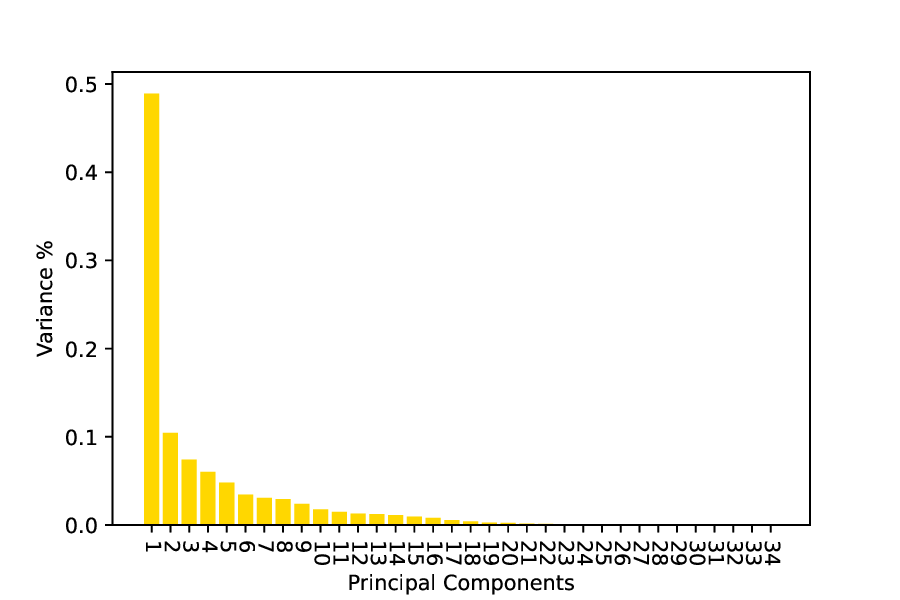}
    \caption{Variance of PCs} 
    \label{fig:varpc}
  \end{subfigure}
  \begin{subfigure}[t]{0.5\textwidth}
    \centering
    \includegraphics[height=3.2in,trim={2.5cm  .1cm 2.5cm 3.9cm},clip]{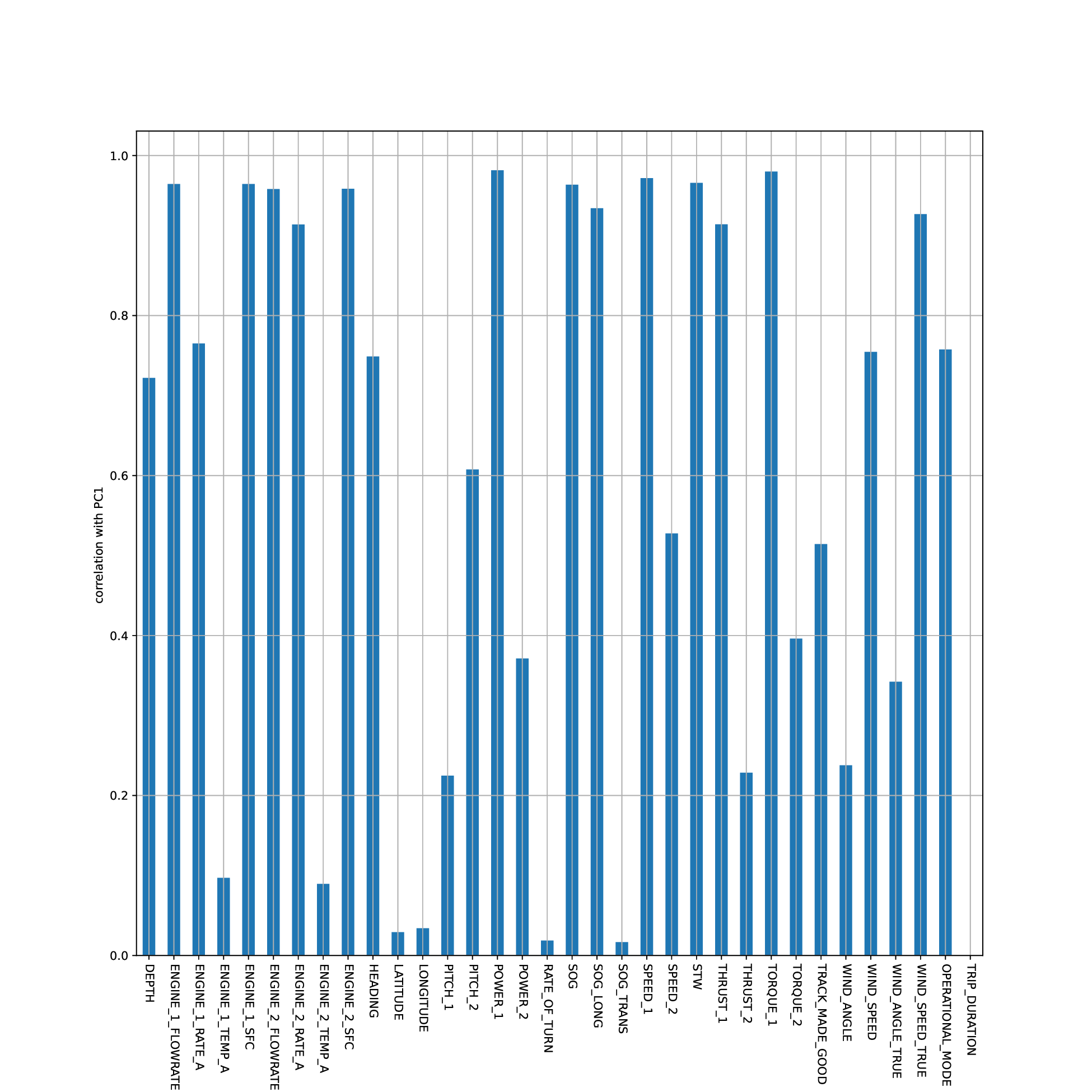} 
    \vspace{0.5ex}
    \caption{Correlation between variables and PC1} 
    \label{fig:corpc}
 \end{subfigure}
  \caption{PC variance and the correlation}
  \end{center}
\end{figure*}

\begin{figure}[t!] 
  \begin{subfigure}{.49\textwidth}
    \centering
    \includegraphics[width=.75\linewidth,trim={.5cm  .1cm 1.5cm 1.2cm},clip]{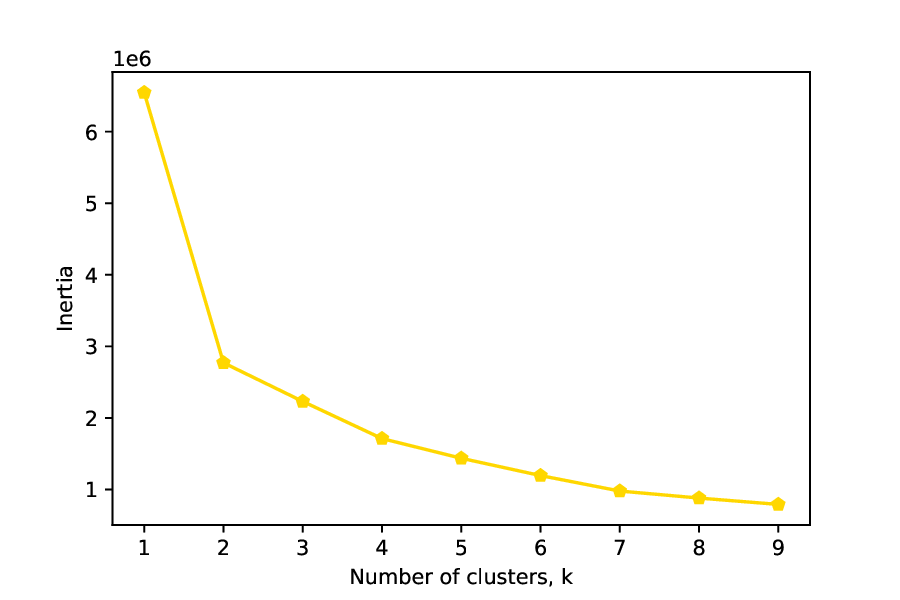} 
    \vspace{1ex}
    \caption{Inertia with respect to number of clusters} 
    \label{fig:inpc}
  \end{subfigure}
  \begin{subfigure}{.49\textwidth}
    \centering
    \includegraphics[width=.8\linewidth,trim={.1cm  .1cm 1.5cm 0.2cm},clip]{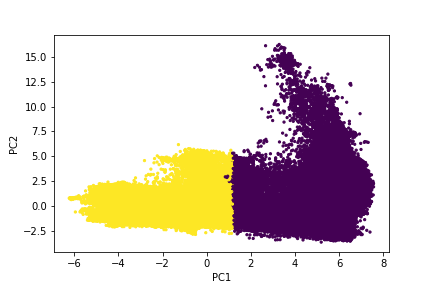} 
    \vspace{1ex}
    \caption{Clusters in PC2 vs. PC1 } 
    \label{fig:clpc}
 \end{subfigure}
 \caption{PC inertia and cluster in PC coordinate}
\end{figure}

PCs are linear combinations of variables that are orthogonal and have no association with each other, with the majority of the data variation condensed into the first few PCs. Fig. \ref{fig:varpc} shows the variance of each PC after applying PCA to our dataset. It can be seen that most of the variance is contained within the first few PCs. Therefore, we only use the first six PCs for K-means clustering. In Fig. \ref{fig:corpc}, the correlation between the first PC (``PC1'') and each variable are plotted as an example. As shown, the correlation to SFC, flow rate, power, and torque are the highest. This can also mean the correlation among themselves is relatively high as their relation to the SFC is approximately linear. \pa{Therefore, we used this knowledge in our feature selection by excluding all of them except the torque, as the latter can show further insight about the applied resistive force to the ship. The detailed information about the dataset variables is shown in appendix \ref{sec:appendixa}}. To choose the number of clusters, we plotted the inertia as defined in Eq. (\ref{eq:inertia}) in Fig. \ref{fig:inpc}. In this case, the elbow occurs after two clusters, so we proceeded with that.

\pa{A projection of the two clusters is shown in Fig. \ref{fig:clpc} in the PC1-PC2 space. We color-coded different clusters from the K-means algorithm in which the yellow represents cluster 1 and purple represents cluster 2.} As we can see, the data is separated into two distinct parts. To get further insight into what these clusters are, we manually partitioned the data according to the operational mode, engine power, and location of the ship. Then, we compared these manually-partitioned clusters with the two obtained using K-means. \pa{The variance of the manually-partitioned cluster was 0.2498, while the mean absolute error between it and the K-means cluster was 0.1343 with a variance of 0.1162. Since the variance was bigger than the variance of their difference, this suggests that the data is mostly clustered based on the operational mode}. The clustering also helps us in outlier detection for steady-state functioning of engine power, while other approaches (e.g., \cite{tsitsilonis2018novel}) have to detect transient data by removing the engine power that fluctuates by more than 5 percent hourly, which is not robust to changes in weather conditions. Therefore, with clustering, we can provide a systematic approach to eliminate the effect of factors such as operational mode, captain, and direction of movement, which are not presented as a separate feature in our dataset while they have an effect on the target variable. At the end, we chose the Mode 1 cluster as it includes around 80\% of the transit time.

\subsection{ Input/target feature selection/engineering}
\pa{For feature selection, we combine correlation evaluation and domain knowledge. Using domain knowledge, we can apply transformations to engineer new features that can reflect the information contained in the dataset more effectively. To obtain further information on the type of relation between fuel consumption, engine speed, and STW, we used the off-diagonal scatter plots in Fig. \ref{fig:hist}. As it can be seen, the histogram of the fuel consumption in the top-left plot shows roughly normally distributed data with low variance, yet for other diagonal plots, the outliers can be detected. The relation with speed is an order 2 polynomial. The scatter plot of STW (left bottom) shows a linear relation, it is sparse, which means the relation is not always valid.}

Assuming similar ship operating conditions, SFC per unit hour is a useful indicator of fuel economy and is a dependent variable in numerous past studies \cite{farag2020development, gkerekos2019machine}. Direct comparisons of fuel consumption per hour, however, may not always be useful due to the fact that fuel consumption and sailing distance rely on a number of variables, including the load of the vessel, sea state, and weather in the navigation zone. Given that the aim of this study is to develop a prediction model to assist ship operators in various navigational circumstances, it is especially recommended that fuel consumption per unit \textit{distance} be used as a measure of ship fuel efficiency. Therefore, the target variable is considered Ship Fuel Efficiency (SFE), instead of SFC, which can be computed via Eq. (\ref{eq:01}). $SFC_1$ and $SFC_2$ refer to the first and second engines' fuel consumption. \rvv{SFE is computed by dividing SFC by travel distance rather than SOG. The reason is that the signal-to-noise (SNR) ratio for GPS location in travel distance is higher than the speedometer using the Doppler approach. This is because, in the former approach, we divide the time elapsed equally ourselves according to the data's sampling ratio (1 minute).}

\rv{In the correlation evaluation step, we compute the linear correlation matrix of all input variables, including the target variable. We retain only one variable from a set of highly correlated variables and sort the remaining features based on their correlation with the target variable. Variables that have a correlation value less than 0.5 are excluded. Subsequently, we employ domain knowledge to engineer new variables and create additional features, such as accounting for the effect of water currents, computing the effective wind, and computing averages for variables that involve two quantities, such as pitch, engine speed, and torque. The list of finalized features is presented in Table \ref{tab:features}.}

\begin{equation}
SFE = \frac{SFC_1 +SFC_2 ~~(\textit{kg per unit hour})}{2\Delta d ~~(\textit{traveled distance per unit hour})}
    \label{eq:01}
\end{equation}


\begin{table}
\caption{List of features and target variable with their description}
\begin{tabular*}{\tblwidth}{L@{}L@{}L@{}L@{}L@{}}
\toprule
 Name &Variable &Unit & Type & Description\\ 
\midrule
Mean pitch & $\bar{P}$&$\%$ & calculated & $(pitch_1 + pitch_2)/2$\\
Engine mean speed &$\bar{V_e}$&RPM&calculated&$(Engine\_speed_1 + Engine\_speed_2)/2$\\
 Speed through water& $STW$ &Knot& measured & relative speed of vessel to water\\
Mean torque & $\hat{\tau}$&kNm& calculated& $(torque_1 + torque_2)/2$\\
Wind angle&$\alpha_w$ &deg &measured& absolute direction of wind\\
Headwind speed& $V_{w,eff}$&knot& calculated& $cos(\alpha_w - \theta)V_w$ $^1$\\
 Heading&$\theta$ &deg & measured &absolute vessel deviation \\
Speed over ground&$SOG$ &knot& measured & absolute speed of vessel \\
 Water current effect&$SOGmSTW$ &knot & calculated &  $SOG-STW$ \\
 Traveled distance &$d$ &deg & calculated & $\sqrt{(\Delta Longitude^2 +\Delta Latitude^2)}$\\
 Fuel efficiency &$SFE$ &kg/deg & Target & $(SFC_1+SFC_2)/(2d)$\\
\bottomrule
$^1$ $V_w$ is the measured apparent wind speed (knot). 
\end{tabular*}
\label{tab:features}
\end{table}

\subsection{Outlier detection}
\begin{figure}[t!] 
  \begin{subfigure}{.49\textwidth}
    \centering
    \centerline{\includegraphics[width=0.85\linewidth,trim={0cm  0cm 0cm 0cm },clip]{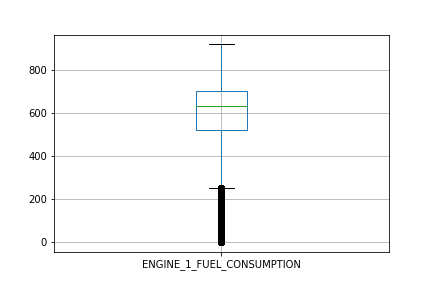}}
\caption{SFC box plot }
    \label{fig:outlier}
  \end{subfigure}
  \begin{subfigure}{.49\textwidth}
    \centering
    \centerline{\includegraphics[width=0.95\linewidth,trim={1cm  0cm 1cm 0cm },clip]{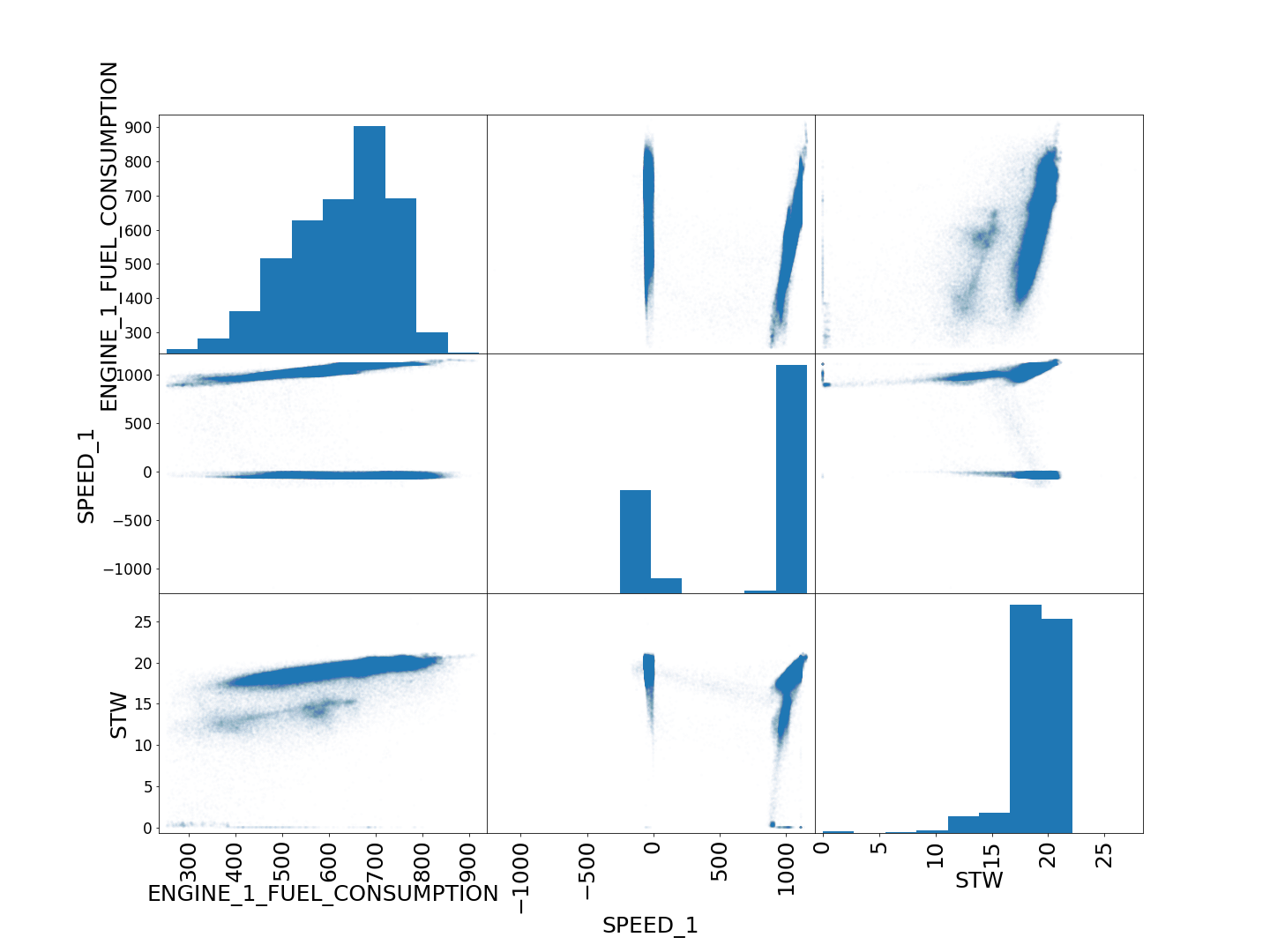}}
\caption{Histogram (diagonal) and scatter plots (off-diagonal ) plots }
    \label{fig:hist}
 \end{subfigure}
 \caption{(a) Outlier detection of SFC based on 1.5 IQR rule and (b)  Histogram and cross correlation scatter plots of SFC, engine speed, and STW}
\end{figure}

\pa{After selecting features, we need to drop non-informative entries (data points) to have a dense distribution. In the outlier detection step, we chose a range such that the original data distribution in that range has relatively low variance. This can be done by using the diagonal plots in Fig. \ref{fig:hist} for SFC and other main features.} Outlier detection is also known as anomaly detection, noise detection, deviation detection, or exception mining. It is defined by \cite{barnett1984outliers} as ``an observation which appears to be inconsistent with the remainder of that set of data''. To help the regression model generalize on unseen data, it is good practice to remove extreme samples from the training dataset that are generated from engine transients and abnormalities in the recording for electronically controlled engines. For instance, the minimum engine speed for continuous operation is between 15 and 20 percent of the engine's nominal maximum continuous speed, and between 20 and 25 percent for camshaft-controlled engines \cite{gkerekos2020novel}. Any observations corresponding to measured engine speed below that threshold are then rejected as an engine in the transient or maneuvering state.


The interquartile range (IQR) is a data-driven approach for outlier detection. The five-number summary can be used to summarize any set of data. The dataset's lowest or minimum value one-quarter of the way through the list of all data is represented by the first quartile Q1. The dataset's median, which is the midway of the entire list of data, three-quarters of the way through the list of all data, or the third quartile Q3, the dataset's highest or highest value $IQR = Q3 - Q1$. We applied the $1.5 \times IQR$ rule to isolate data lower than 1.5 times the IQR under the first quartile and higher than 1.5 times above the third quartile. Note that this is not necessary for decision tree-based predictors, which are robust to outliers. For instance, the outlier detection for fuel consumption is shown in Fig. \ref{fig:outlier}.

\subsection{Handling missing data}
Common approaches to handle missing data are to either drop all the entries of missing data, set missing data to zero, or impute the missing values with the most common value or the median value. Our dataset was not different. The missing data was imputed for entries with only one variable missing by within-cluster mode. Otherwise, we deleted the entry.

\subsection{Dataset split}
\rv{To mitigate the risk of overfitting, the dataset was randomly partitioned into training and testing sets with a ratio of 70:30, respectively. The training set was utilized for model training, while the testing set was reserved solely for assessing the final performance metrics. Furthermore, to fine-tune hyperparameters and evaluate model performance, K-fold cross-validation was employed during the training process. Specifically, one-tenth of the data was randomly selected for each iteration of cross-validation \cite{wong2019reliable}}.

\section{Modeling}
In this section, statistical approaches are first used to gain insight about the data. In the following section, we present ML techniques to find a prediction model for fuel consumption given the selected features and the preprocessed data from the previous section. Finally, we compare the different ML approaches for the dataset in terms of running time, accuracy, and prediction capability.

\subsection{Statistical study}
\begin{figure}[t!]
\begin{center}
\centerline{\includegraphics[width=.8\linewidth,trim={1.5cm  15cm 1cm 4cm },clip]{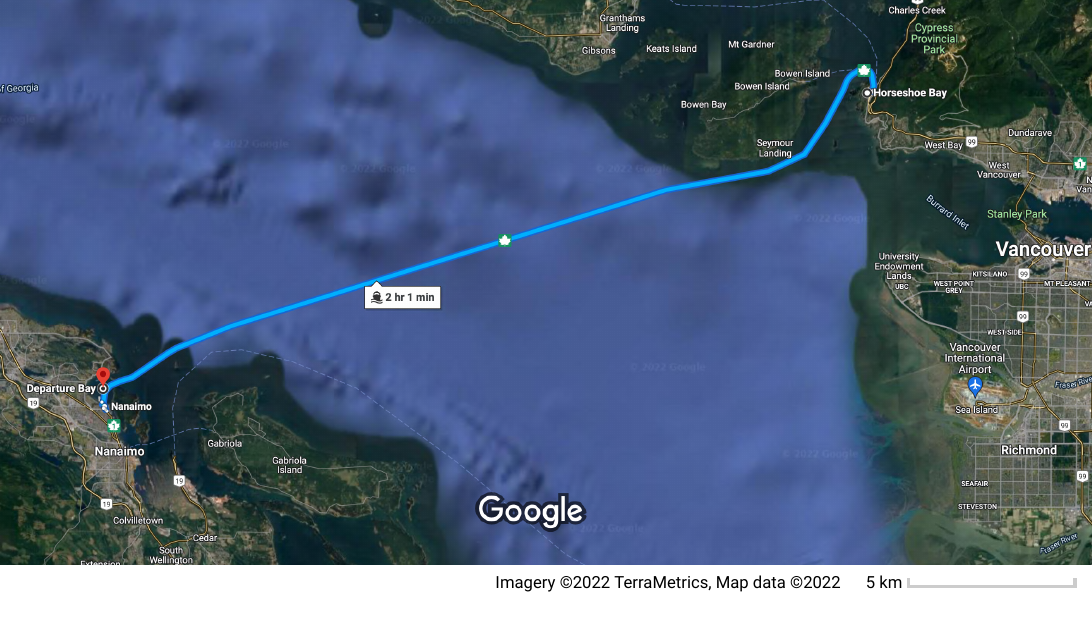}}
\caption{Google map of the transit from two docking points}
 \label{fig:map}
\end{center}
\vskip -0.2in
\end{figure}

\begin{figure}[t] 
  \begin{subfigure}{.49\textwidth}
    \centering
    \includegraphics[width=0.9\linewidth]{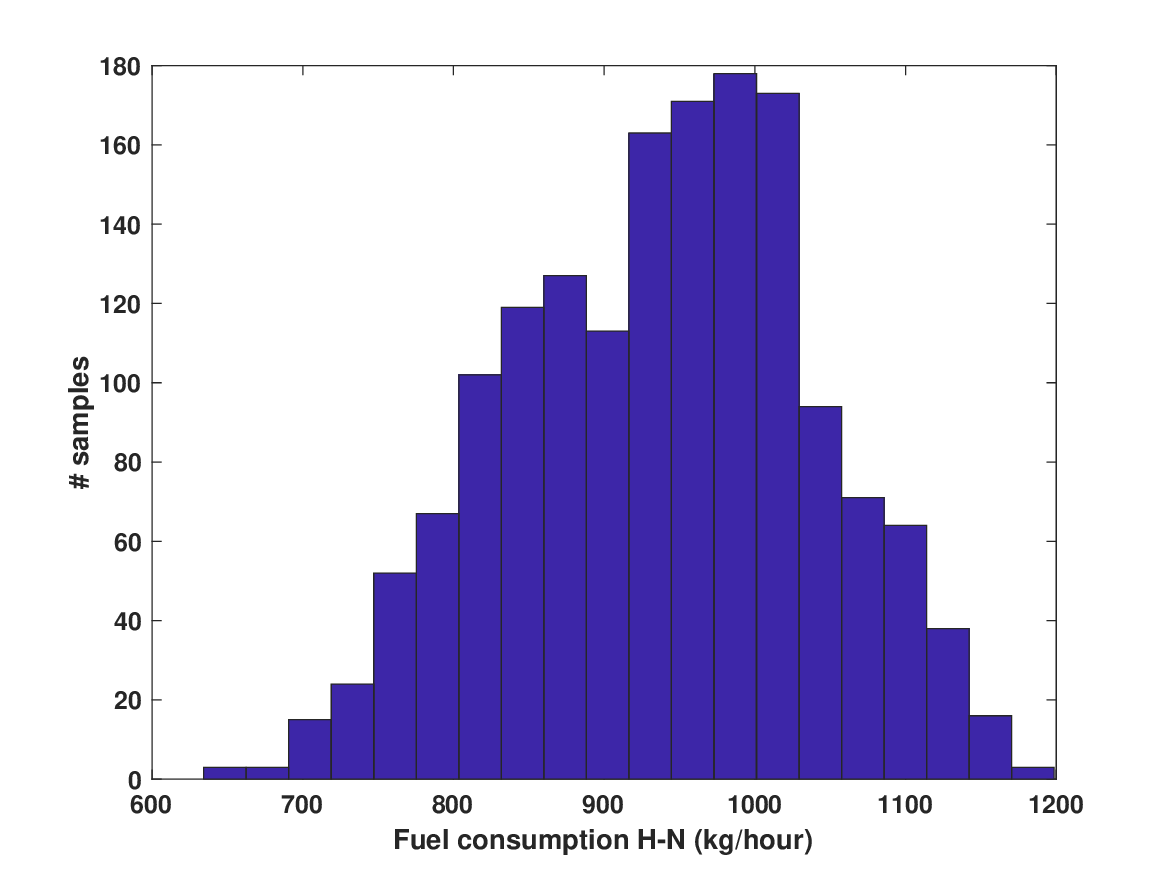}
    \vspace{1ex}
    \caption{H-N trip}
  \end{subfigure}
  \begin{subfigure}{.49\textwidth}
    \centering
    \includegraphics[width=0.9\linewidth]{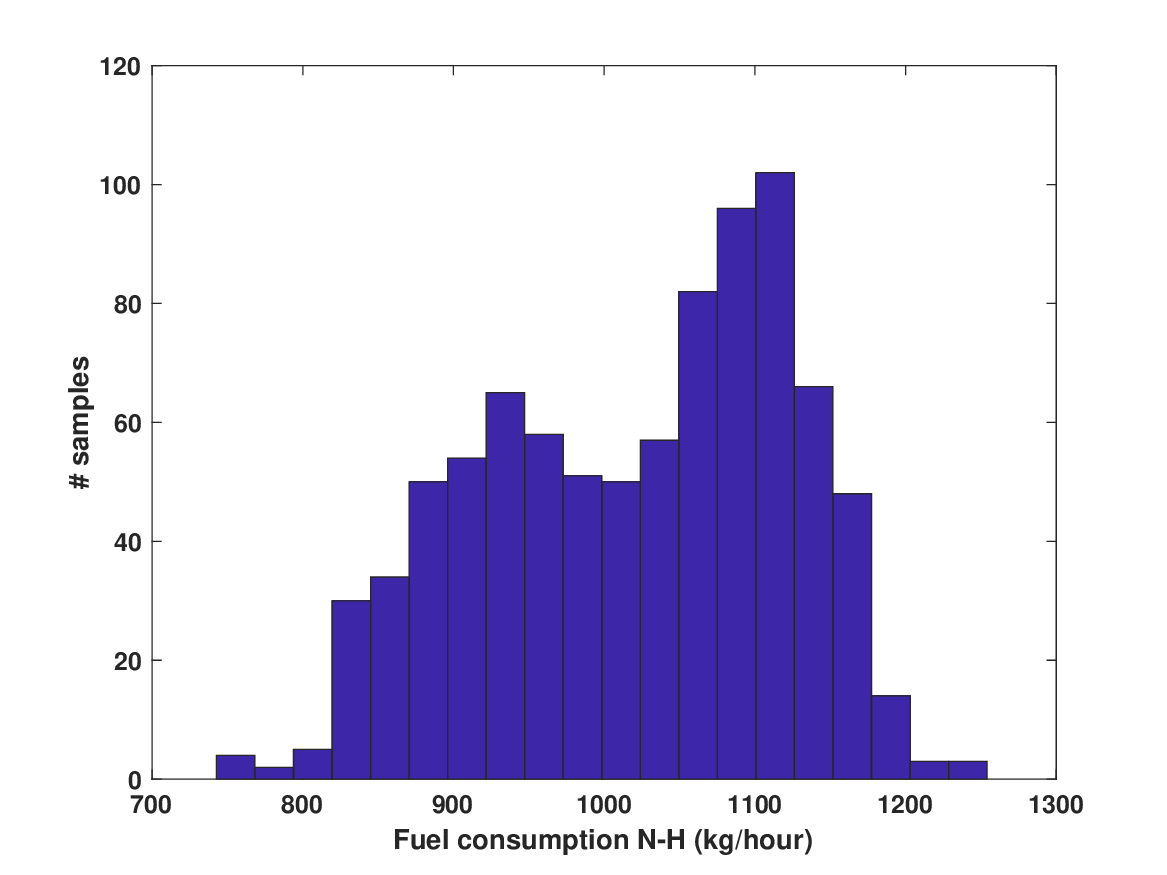}
    \vspace{1ex}
    \caption{N-H trip}
 \end{subfigure}
  \caption{\pa{SFC (kg/hour) histogram for sample trips: (a) H-N and (b) N-H}}
  \label{fig:fc}
\end{figure}

Fig. \ref{fig:map} shows a section of the map that contains the two docking points of the target ship: Horseshoe Bay and Nanaimo. The statistical analysis of sample routes from Horseshoe Bay to Nanaimo (H-N) and Nanaimo to Horseshoe Bay (N-H) is illustrated in Table \ref{tab:stat}. In this table, the columns represent the best and second-best trips over two years of our data collection with respect to the SFC, along with the average trips of H-N and N-H. $\theta_\delta$ is the heading deviation from the straight line that connects the two docking points. The histogram for SFC of the first engine for each route is shown in Fig. \ref{fig:fc}. \pa{There are mostly six trips back and forth each day between H-N and N-H. As we can see, in terms of fuel consumption, the H-N and N-H trips are not symmetric, as the average of H-N SFC is lower. In addition, due to the usual west-to-east wind, the H-N trips are more against the wind. The large difference between the best and worst trips suggests that using improvements in navigational practices may result in fuel consumption savings.}

\begin{figure}[t] 
  \begin{subfigure}{.49\textwidth}
    \centering
    \includegraphics[width=0.9\linewidth]{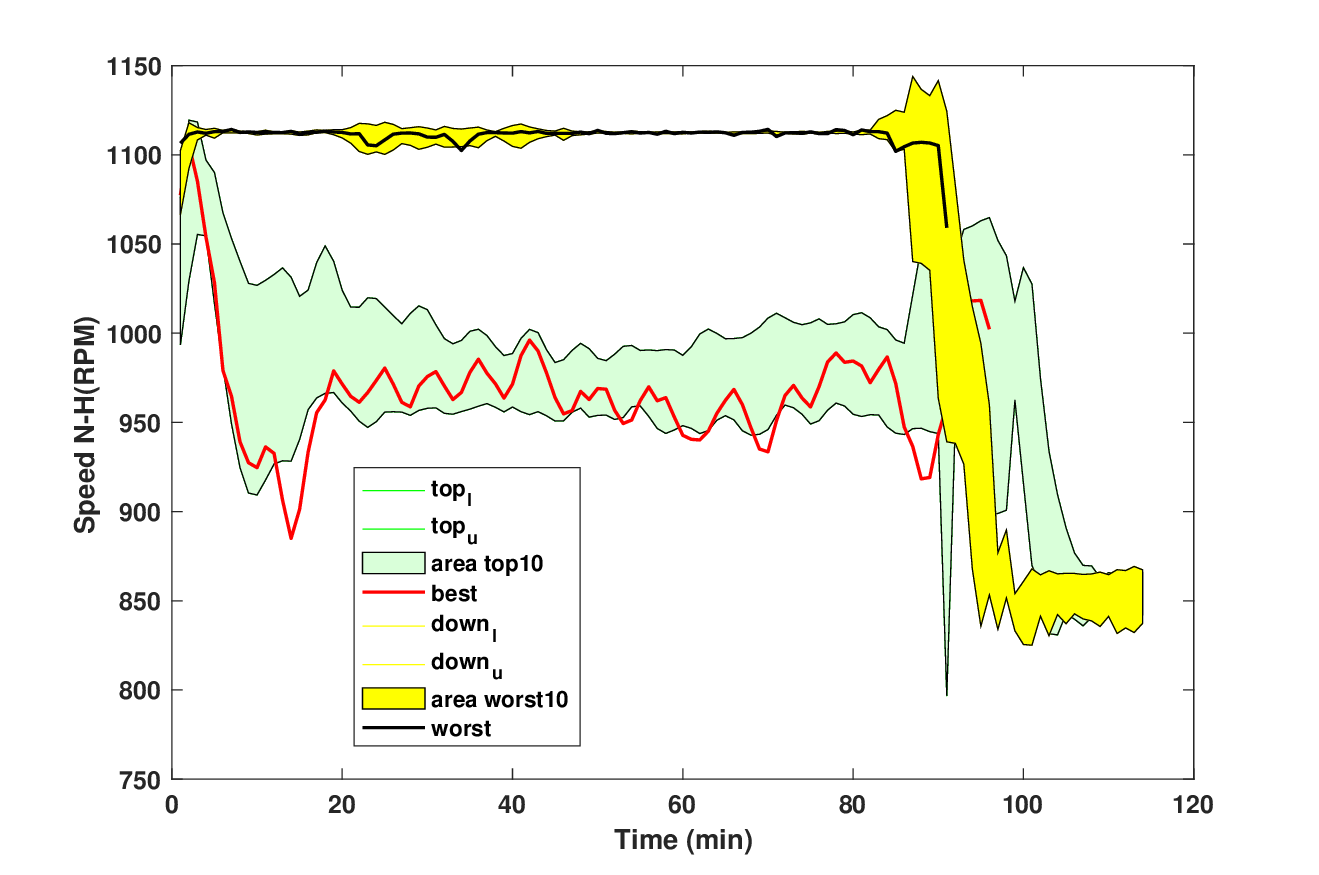} 
    \vspace{1ex}
    \caption{Engine speed (RPM)}
  \end{subfigure}
  \begin{subfigure}{.49\textwidth}
    \centering
    \includegraphics[width=0.9\linewidth]{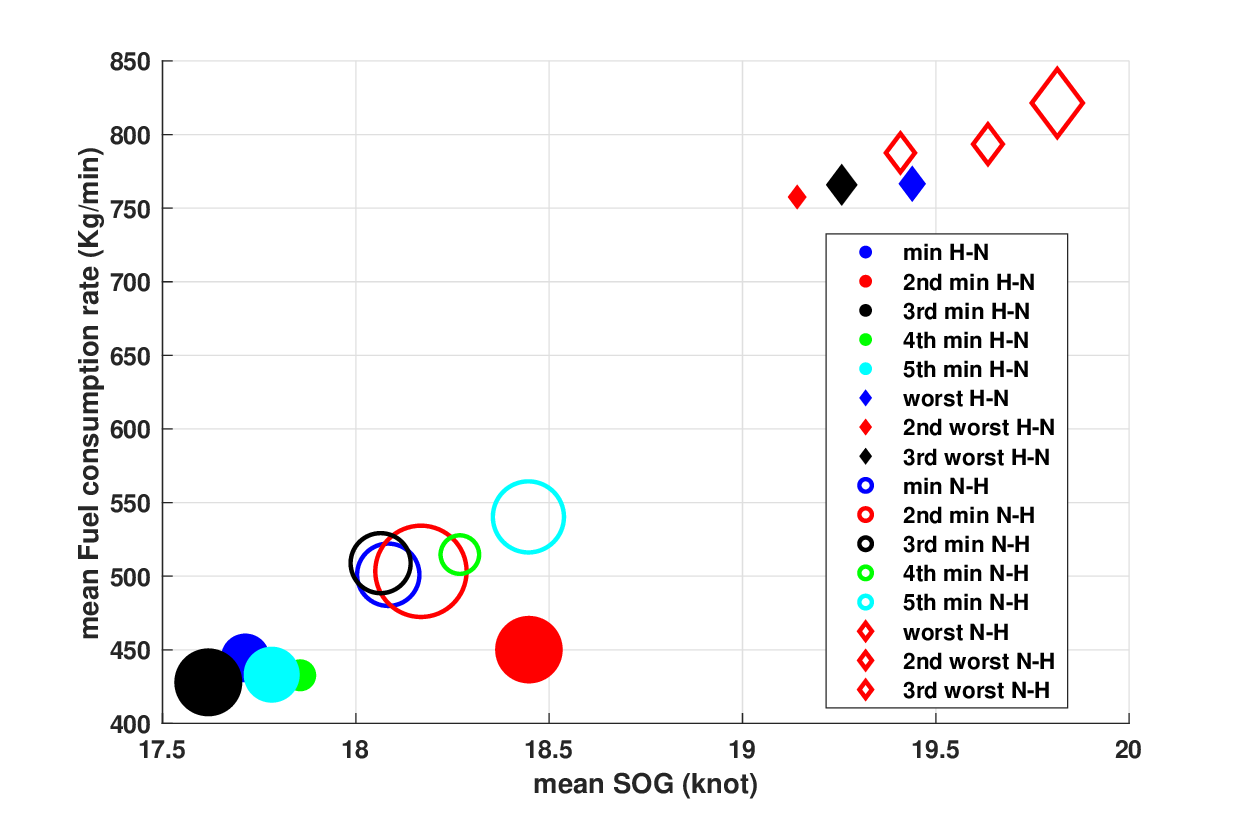}
    \vspace{1ex}
    \caption{ SFC (kg/hour) vs SOG (knot)}
 \end{subfigure}
  \caption{\pa{(a) Engine speed (RPM). ``Top'' refers to the top 10\% quartile
 and ``down'' refers to the worst 10\% quartile in SFC. Subscripts $u,l$ denotes the upper and lower bound, respectively. (b) SFC  vs SOG for different trips. The size of the shapes corresponds to the variance of the SFC throughout that trip.}}
  \label{fig:speed}
\end{figure}

\begin{figure}[t] 
  \begin{subfigure}{.49\textwidth}
    \centering
    \includegraphics[width=0.85\linewidth]{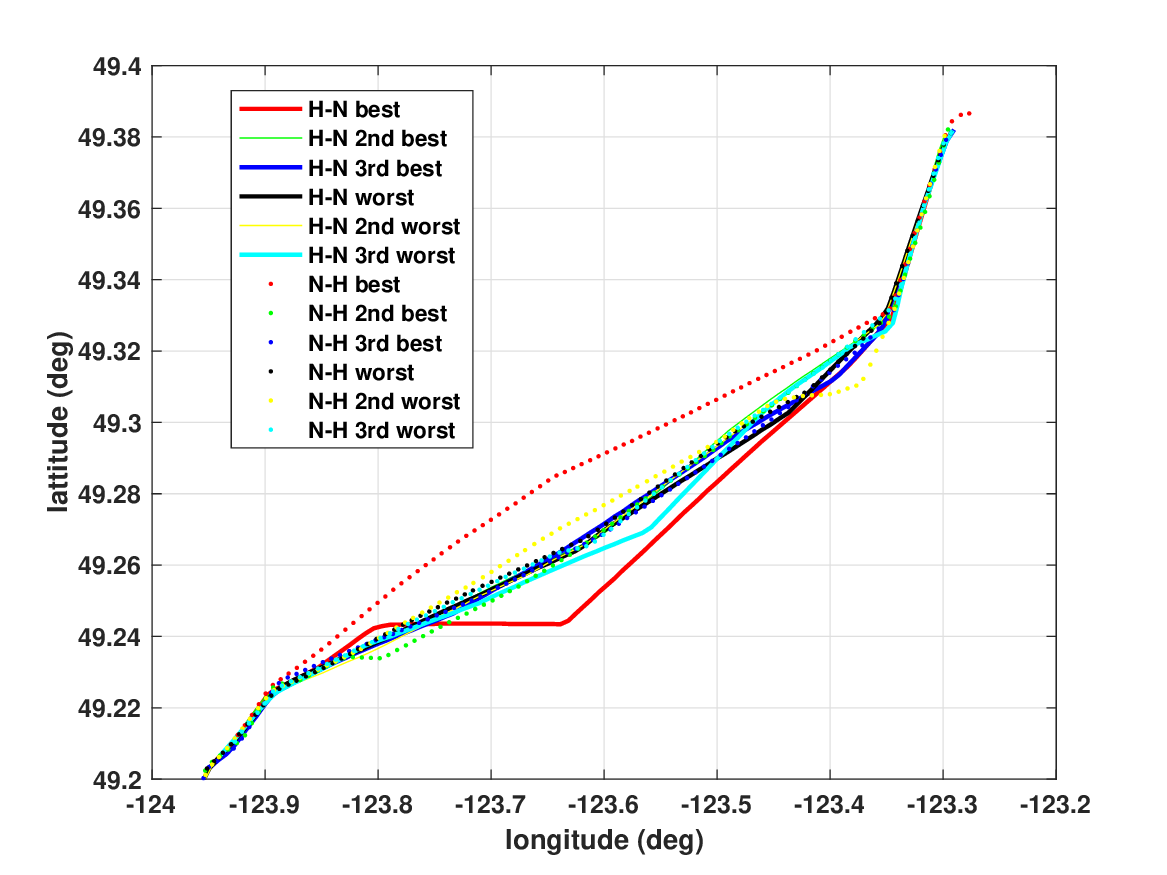} 
    \vspace{1ex}
    \caption{Map of the traveled path}
  \end{subfigure}
  \begin{subfigure}{.49\textwidth}
    \centering
    \includegraphics[width=1\linewidth]{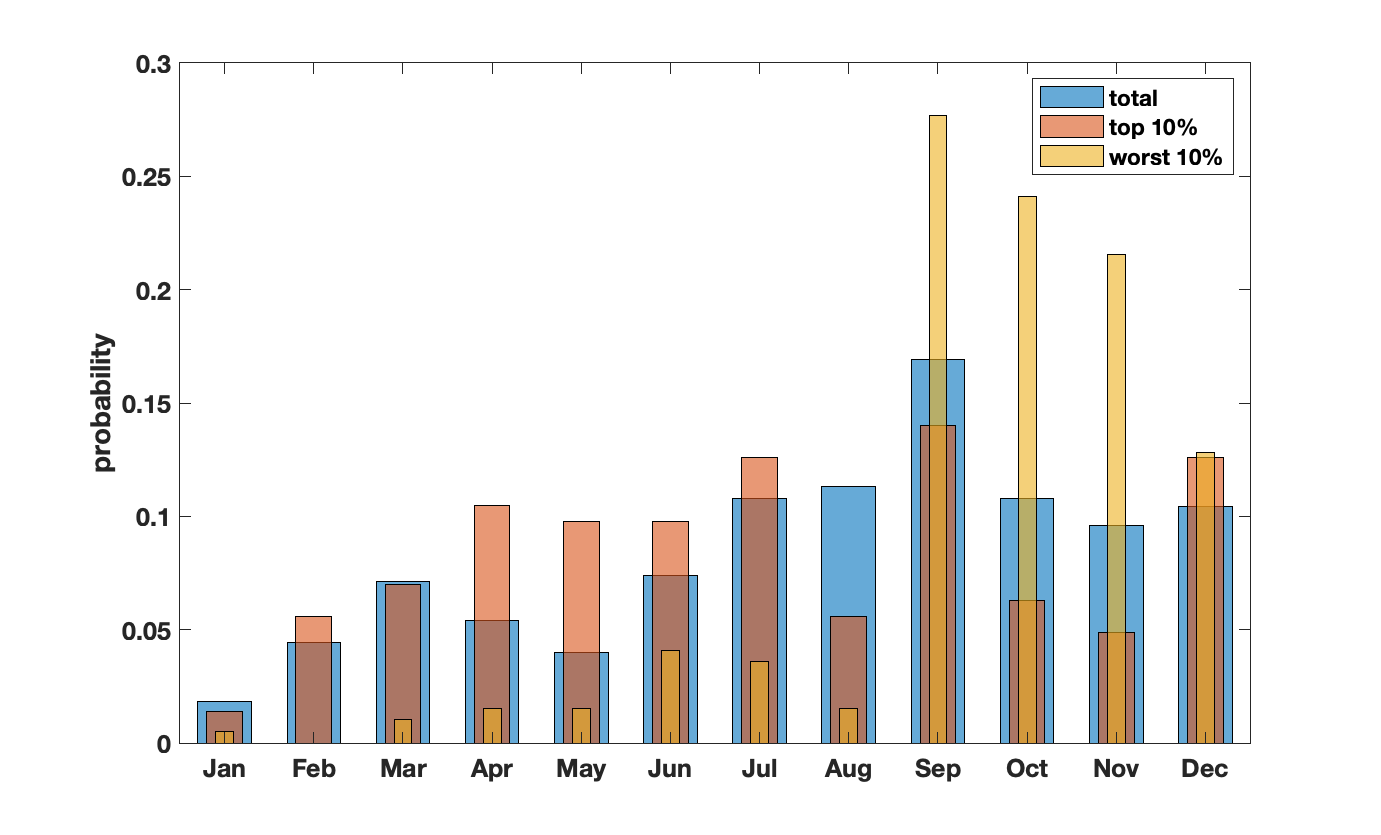}
    \vspace{1ex}
    \caption{Profile of SFC trips for different months}
 \end{subfigure}
  \caption{\pa{(a) The traveled path for different transits in latitude and longitude scale (b) average SFC profile for different months of the year}}
  \label{fig:gps}
\end{figure}

The statistical analysis of the input features after the preprocessing stages is shown in Table \ref{tab:featureanalysis}. \rv{We compute the cross-correlation matrix for all the variables, including the target. The Mean Absolute Error (MAE) of the elements in the matrix shows collinearity between the input variables, as well as correlation with the target variable \footnote{\rvv{The confusion matrix is available in this link \url{https://github.com/pagand/model\_optimze\_vessel/blob/OE/Feature/PCA_clustering\_revision.ipynb}}}}. This shows the effectiveness of each variable in target prediction. \rv{We exclude redundant variables and ignore those with a non-significant correlation to the target.} In Fig. \ref{fig:speed}, the engine speed profile is illustrated for sample routes beside the SFC profile for different SOG. \rv{The green area in the engine speed profile shows the confidence interval for the top 10\% quantile of the SFC, while the yellow region depicts the worst 10\%. The red and black lines represent two samples in each region, respectively. In the scatter plot, the circular shapes represent the best quantile, and the diamond shapes represent the worst quantile, with their size indicating their variance. We can see that the top quantile is in the lower left corner with mostly larger variance, while the worst quantile is in the upper right corner.} We can conclude that for higher SOG, the SFC is less efficient as it lies towards the upper side of the connecting line of lower SOG. This shows that the efficiency decreases in cases of very high SOG. Finally, in Fig. \ref{fig:gps}, the effect of different months of the year on the SFC can be seen. \pa{Compared to other seasons, the Fall season has the most trips with high SFC}. It also shows the GPS plot of different routes, which shows the extent of diversity in different trips. The changes in the route can be the result of a craft in the way or severe weather conditions.

\begin{table}
\caption{Statistical analysis of dataset for different trips (H-N or N-H) before preprocessing}
\begin{tabular*}{\tblwidth}{@{}LC@{}C@{}C@{}C@{}C@{}C@{}}
\toprule
Metric& H-N best & H-N 2nd best & H-N average&N-H best& N-H 2nd best&N-H average \\ 
\midrule
 Date &20/12/03 &21/09/29 &NA&20/11/22 &20/11/23 &NA\\
 Shift  &morning& noon&NA& afternoon& afternoon&NA\\
 $SFC$ (kg/hour)&703.75&705.15&939.16&801.53&805.28&1.02e+03\\
 $SOG$ (Knot)&17.71&18.44&19.03&18.08&18.16&19.11\\
 $STW$ (Knot)&17.31&18.36&19.01&18.03&18.27&19.40\\
 Trip duration (min)&94&93&90.15&95&95& 89.79\\
 $V_{w,eff}$ (knot)&-2.64&-7.14&-8.86&-8.33& 0.24&-4.94\\
 $\theta_\delta$ (deg)& 2.52&1.84&1.81&1.14&1.06&1.08\\
\bottomrule
\end{tabular*}
\label{tab:stat}
\end{table}

\begin{table}
\caption{Statistical analysis of input variables after preproccessing}
\begin{tabular*}{\tblwidth}{@{}CC@{}C@{}C@{}C@{}C@{}C@{}C@{}}
\toprule
Parameter& Mean& Std. Dev &Min& Max&Median&target correlation&MAE cross-correlation\\ 
\midrule
 $\bar{P}$&229.08&175.33&0&530.90&357.94& 0.0928&0.2735\\
$\bar{V_e}$&538.17&92.11&0&1129.09&526.31&-0.0202&1.1118\\
$STW$ &18.69&2.17&0&27.72&19.26&0.4256&2.9580\\
$\hat{\tau}$&158.72&26.13&-0.55&217.71&163.17&0.8117&1.9479\\
$\alpha_w$&130.10&150.38&0&366.00&29.00&-0.0637&0.4286\\
 $V_{w,eff}$&-3.96&4.16&-10.13&12.17&-4.24&0.2774&1.4333\\
$\theta$ &246.37&22.15&0&359.90&250.90&-0.0846&0.8908\\
$SOG$ &18.62&2.09&0&21.87&19.10&0.3870&2.6873 \\
$SOGmSTW$ &-0.07&0.45&-8.40&14.41&-0.09&-0.2554&1.1083 \\
$d$ &0.0095&0.0360&0&0.7091&0.0077&-0.0643&0.6783\\
$SFE$ &630.52&109.94&253.84&1000.00&651.04&1.0000&NA\\
\bottomrule
\end{tabular*}
\label{tab:featureanalysis}
\end{table}

\subsection{Machine learning approaches }
Based on the features' statistical analysis conducted in Table \ref{tab:featureanalysis}, the features do not have the same range of values. Features with a large range may have a dominating effect on the model when compared to the smaller range input variables. As such, feature re-scaling should be considered. Although this may not raise an error in models that do not take a distance-based approach, such as tree-based models (e.g., DT, RFR, etc.), for other approaches such as deep ANN, it will deteriorate the training time. Data standardization (mapping the data between 0 and 1) and normalization (mapping the data to have a mean equal to zero and variance equal to one) are linear transformations that speed up the training process and avoid computational errors. In this work, we employed the latter data normalization approach.


The criteria for evaluating the prediction models are the root mean square error (RMSE) and $R^2$, defined as follows:

\begin{equation}\begin{split}
RMSE =& \sqrt{\frac{\sum_{i=1}^{N}(y_i - \hat{y}_i)^2}{N}},\\
R^2 = & 1 - \frac{\sum_{i=1}^{N}(y_i - \hat{y}_i)^2}{\sum_{i=1}^{N}(y_i - \bar{y}_i)^2},
\end{split}
\label{eq:eval}
\end{equation}
where $y_i$ is the $i$-th observed target value, $\hat{y}_i$ is the predicted value, and $\bar{y}_i$ is the mean of the observed value.

\begin{figure}[t] 
  \begin{subfigure}{.49\textwidth}
    \centering
    \includegraphics[width=0.85\linewidth]{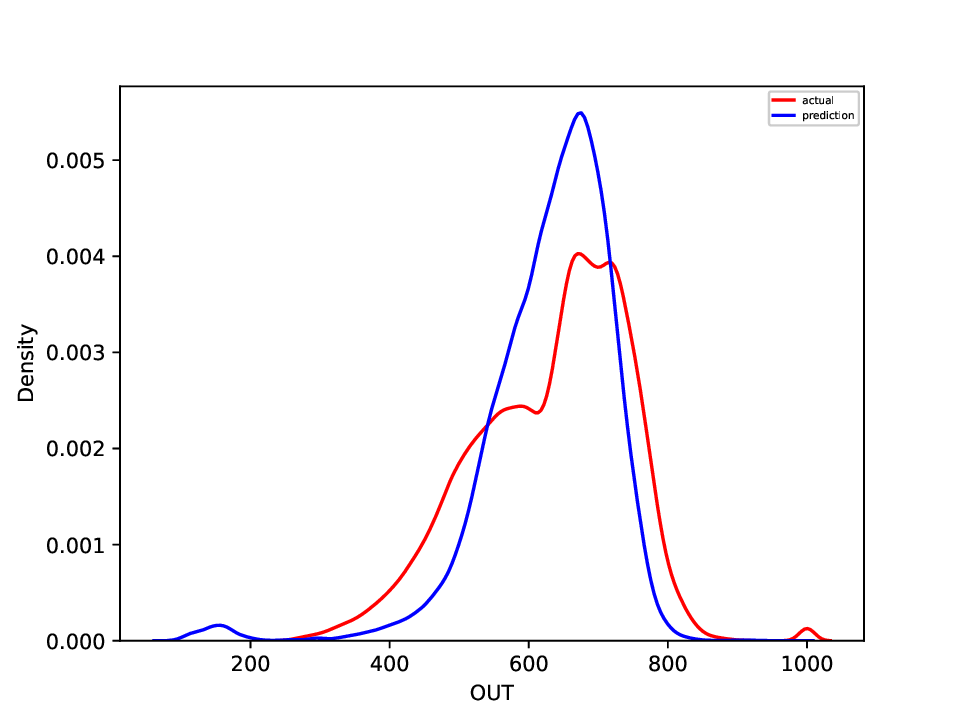} 
     \caption{Linear regression}
    \vspace{1ex}
  \end{subfigure}
  \begin{subfigure}{.49\textwidth}
    \centering
    \includegraphics[width=0.85\linewidth]{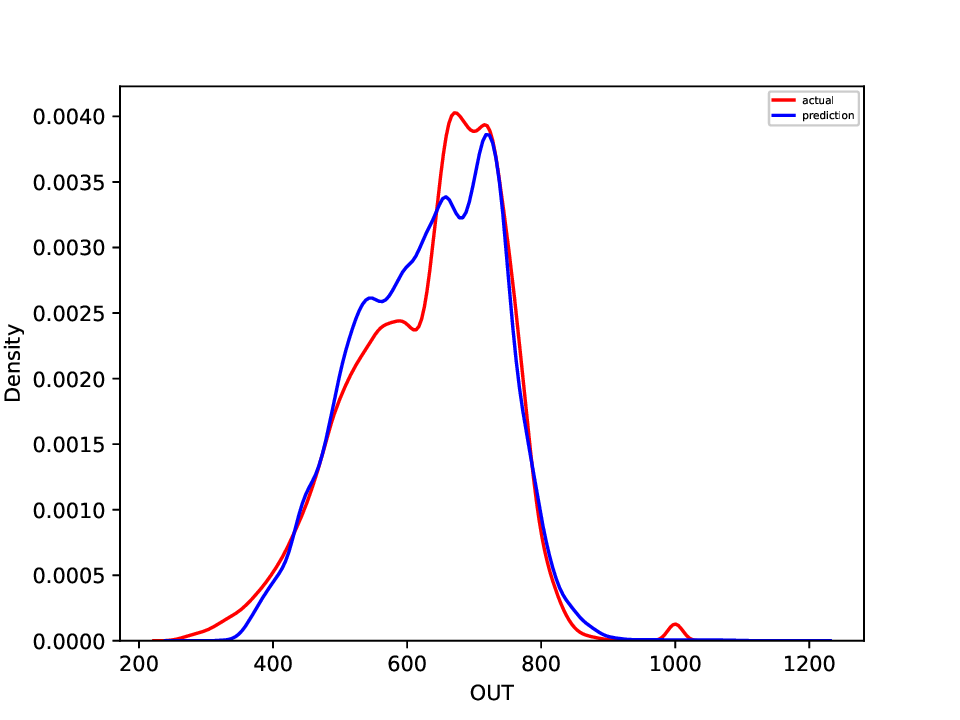}
    \caption{Polynomial regression}
    \vspace{1ex}
 \end{subfigure}
  \caption{Histogram plot of SFC test dataset for linear and polynomial degree 2 regression}
  \label{fig:ml1}
\end{figure}

\begin{figure}[t] 
  \begin{subfigure}{.49\textwidth}
    \centering
    \includegraphics[width=0.85\linewidth]{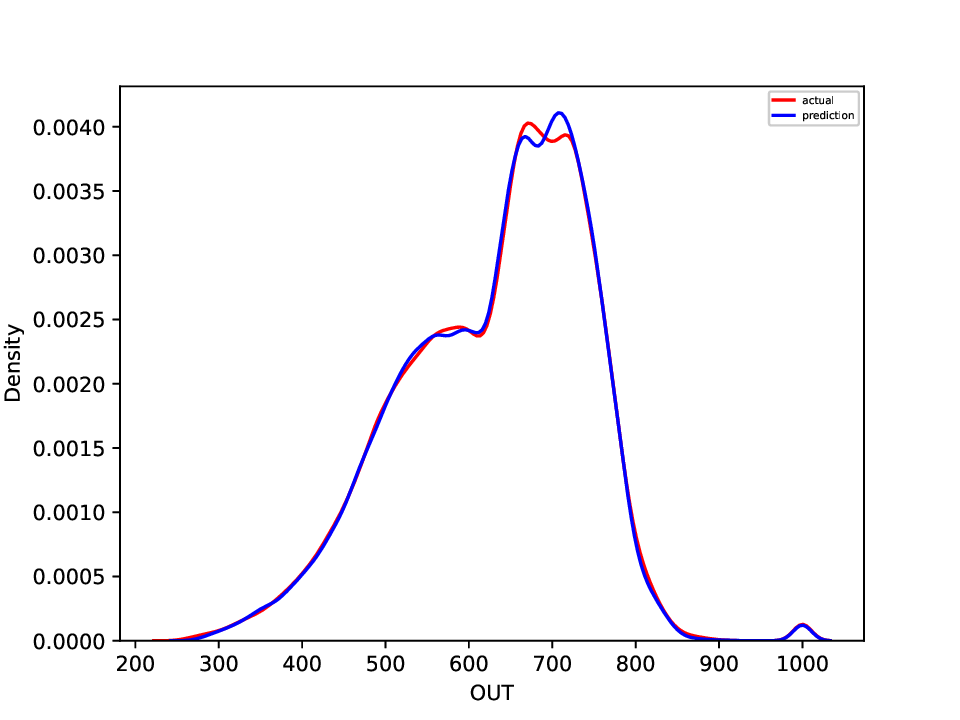} 
     \caption{Random forest regression}
    \vspace{1ex}
  \end{subfigure}
  \begin{subfigure}{.49\textwidth}
    \centering
    \includegraphics[width=0.85\linewidth]{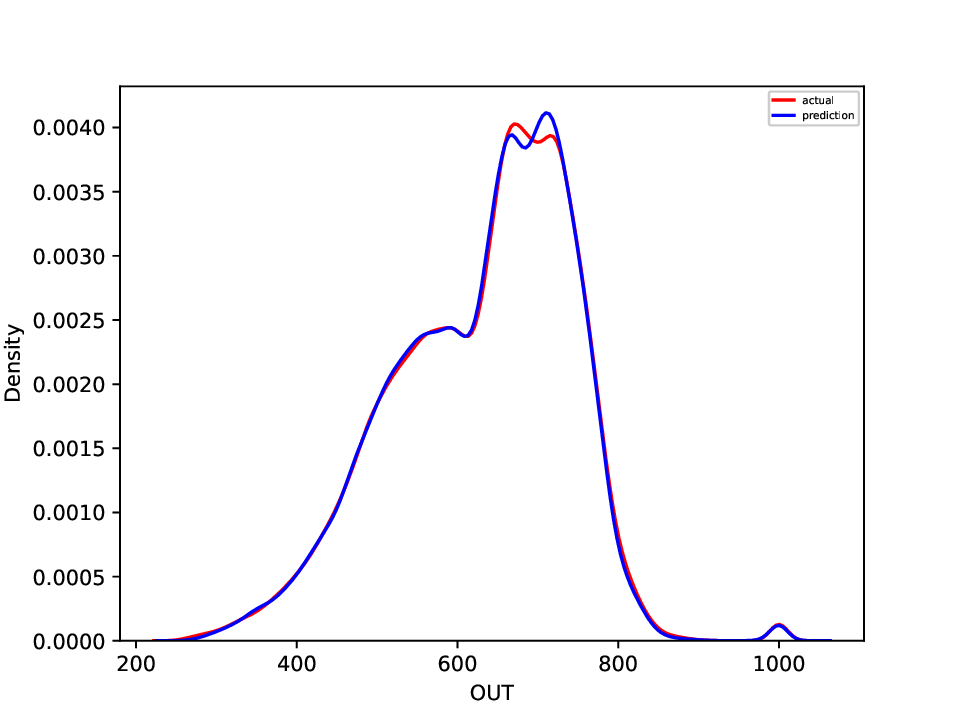}
    \caption{XGBoost regression}
    \vspace{1ex}
 \end{subfigure}
  \caption{Histogram plot of SFC test dataset for decision tree and random forest}
  \label{fig:ml2}
\end{figure}

The results of linear and polynomial regression are shown in Fig. \ref{fig:ml1}. RFR and XGBoost results are shown in Fig. \ref{fig:ml2}.  \pa{These plots illustrate the density of the SFC in different regions. Ideally, the predicted and actual plots should match in all regions, but as seen, there are some regions where the densities are swapped or not exactly matching. This happens due to a mismatch in the prediction results. This mismatch happens for XGBoost in its most frequent region around 700 $kg/h$.}

\subsubsection{Hyperparameter tuning}
For implementation, an NVIDIA RTX 3080 Ti GPU, PyTorch v1.8.1., scikit-learn v1.1.2, and Python v3.7.4 were employed. For Lasso, the $\alpha$ parameter was set to 0.1. During training, the parameter STW and DISP shrink to zero, meaning these parameters are the least important parameters in the regression model. For Ridge regression, the $\lambda_1$ parameter was set to 10. For DT and all ensemble techniques, a randomized K-fold cross-validated search over parameters was used to find the suitable hyperparameters. Unlike grid-based cross-validation, not all hyperparameter values are tried out, but rather a fixed number of hyperparameters are sampled from the specified distributions. The final hyperparameters are shown in Table \ref{tab:ensemble}. We have tested ``linear'', ``square'', and ``exponential'' for AdaBoost loss function, and ``exponential'' showed the best performance. For gradient boosting, we compared 0.5 and 1 for subsample, and 0.5 performed better. For XGBoost, we compared ``min child weight'' in [1, 3, 5, 7], $\gamma \in [0.0, 0.1, 0.2, 0.3, 0.4]$, and ``colsample bytree'' in [0.3, 0.4, 0.5, 0.7], where 1, 0.2, and 0.7 were the optimal values, respectively. For MLP, we considered 1000 epochs and three hidden layers. The number of neurons in each layer was 30, 20, and 20 for the hidden 1, hidden 2, and output layer, respectively.

\begin{table}
\caption{Ensemble techniques optimal hyper-parameters}
\begin{tabular*}{\tblwidth}{@{}LC@{}C@{}C@{}C@{}C@{}}
\toprule
 Parameter & DT&RF&Ada-Boost& Gradient boosting& XGBoost\\ 
\midrule
Max depth&30&25&40&35&35\\
Max feature&'auto'&'auto'&'auto'&'sqrt'&NA\\
Min samples leaf&12&6&2&12&NA\\
Min samples split&40&40&40&10&NA\\
learning rate&NA&NA&0.3&0.1&0.1\\
\multirow{3}{*}{others}&\multirow{3}{*}{NA}&\multirow{3}{*}{NA}&\multirow{3}{*}{loss:exponential}&\multirow{3}{*}{subsample:0.5}&min child weight:1\\
&&&&&$\gamma$:0.2\\
&&&&&colsample bytree:0.7\\
\bottomrule
\end{tabular*}
\label{tab:ensemble}
\end{table}

\subsubsection{Comparison results}
\rv{Table \ref{tab:quant} presents a quantitative comparison of the different modeling approaches used in the study. The parametric models, except for ANN, exhibit relatively faster training times, while non-parametric approaches result in lower test and validation errors. Among the ensemble techniques, XGBoost has the shortest training time and the lowest test error. By sliding the target variable to train/test for future steps, parametric approaches provide more robust performance compared to non-parametric approaches. While ANN exhibits the largest training time and outperforms other parametric approaches in terms of test results, its performance is still inferior to that of ensemble techniques.}

\rv{The performance of LR, Lasso, and Ridge models is found to be limited, while PR results in significant improvement. DT performs similarly to PR in the non-parametric approach. However, the ensemble techniques demonstrate superior performance compared to parametric and non-parametric approaches, with XGBoost being the best performer. Comparing the different parameter future prediction errors, MLP shows the greatest deterioration in performance compared to other approaches due to its inability to capture the dynamic behavior of the system. Although XGBoost's performance also degrades for a 5-step ahead prediction, it still provides the best estimation among all approaches.}

\rv{Comparing the train and test running times, conventional parametric and non-parametric approaches are relatively faster than ensemble techniques and NN approach. Among ensemble techniques, XGBoost provides results more than twice as fast compared to AdaBoost/gradient boosting, which is important for real-time implementation. Additionally, comparing the training times allows for a fair comparison. For example, RF exhibits a validation error that is 40\% better than LR but is more than 800 times slower.}

\begin{figure}[t] 
    \centering
    \includegraphics[width=0.75\linewidth]{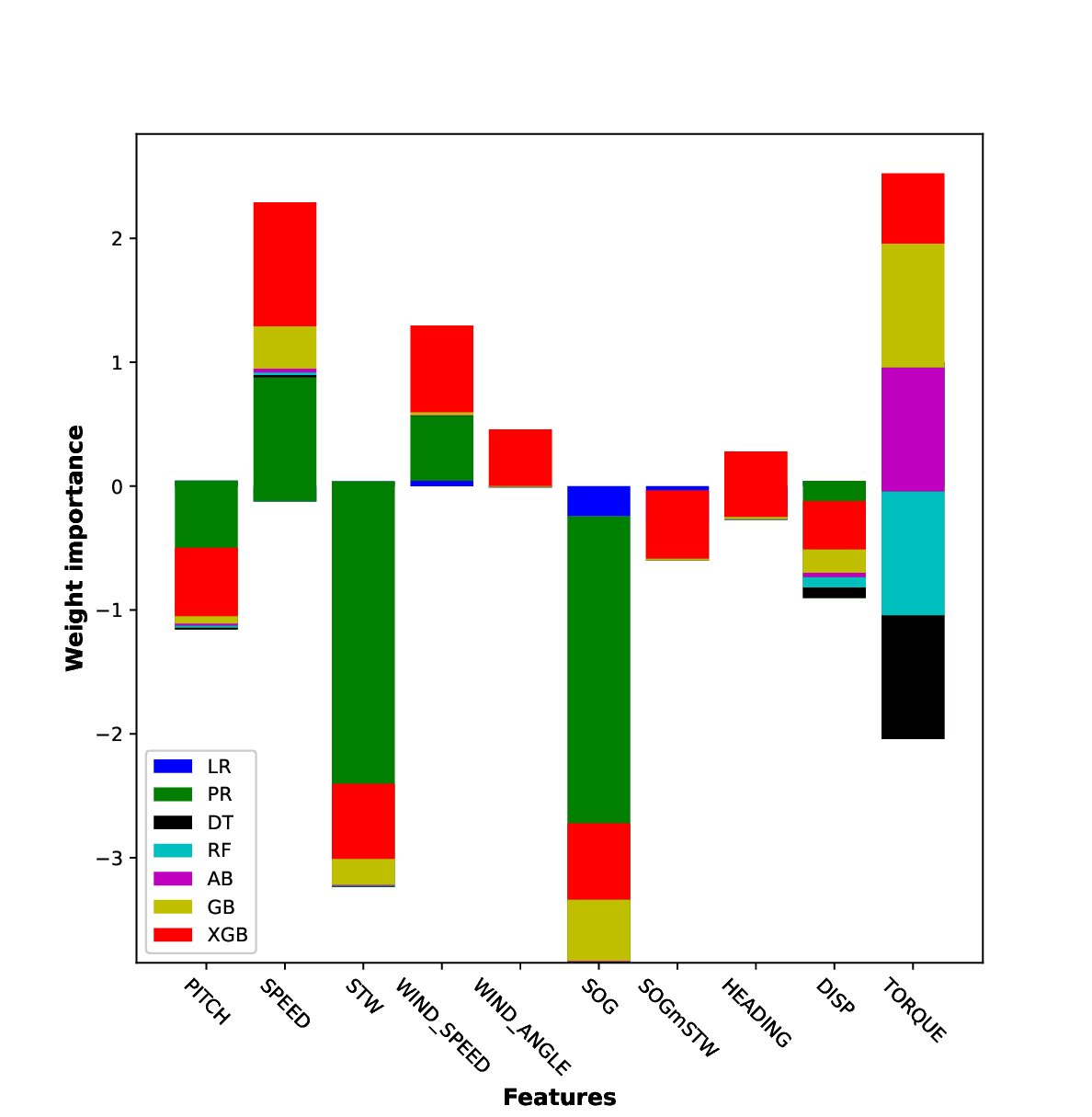}
    \caption{Comparison for feature importance  for different techniques }
  \label{fig:fi}
\end{figure}

The importance of different features in the approaches is compared in Fig. \ref{fig:fi}. As it can be seen, the torque, SOG, STW, and engine speed are the top four important features as they have the highest length among all approaches, while the wind angle, SOGmSTW ($SOG-STW$), and heading are the least important features. The second takeaway is that XGBoost considered all the features with approximately the same importance. This means that it is the most compliant method to our physical white box model. Our third takeaway point is regarding the DT or AdaBoost method, which almost only depends on the torque and displacement. In some applications, this can be considered as a positive point, since the performance almost only depends on the precision of these two features. In other words, for cases where the best-performed method is not desired or not all features are available, these methods are suggested as alternatives. However, in some cases where the vessel is tested outside the training region or when the disturbances such as the wind or waves significantly affect its normal distribution, the lack of extrapolation in these methods will deteriorate their performance.

\begin{table}
\caption{Quantitative analysis of different ML approaches}
\begin{tabular*}{\tblwidth}{@{}LC@{}C@{}C@{}C@{}C@{}C@{}}
\toprule
 Method & train/test time&\# parameters&$R^2$ validation& $R^2$ Test & RMSE test& future RMSE test\\ 
\midrule
Linear regression &0.40129/0.00096&10& 0.6999&0.6930&60.86108&90.84561\\
Lasso&0.197186/\textbf{0.00104}&\textbf{8}&0.6973&0.6902&61.13519&91.04635\\
Ridge&\textbf{0.14173}/0.00212&10&0.6999&0.6929&60.86271&90.85122\\
Poly regression&0.792789/0.01818&66&0.9306&0.9321&28.60731&77.28871\\
Decision tree&5.87621/0.01330&NA&0.9787&0.9799&15.55609&70.92789\\
Random forest&339.65/1.15040&NA&0.9836&0.9845&13.63560&64.02640\\
Ada-Boost&676.59/1.764294&NA&0.9860&0.9871&12.44784&\textbf{63.65051}\\
Gradient boosting&417.85/1.106061&NA&0.9867&0.9876&12.20962&63.94911\\
XGBoost&283.49/0.31252&NA&\textbf{0.9874}&\textbf{0.9881}&\textbf{11.94711}&67.39446\\
Multilayer perceptrons&7166.24/0.50567&971& 0.9779&0.8327&44.96602&89.49549\\
\bottomrule
\end{tabular*}
\label{tab:quant}
\end{table}
\section{Conclusion}\label{conclusion}
\rv{In this study, we present an end-to-end approach for predicting SFE for a passenger vessel using a gray-box approach. The analysis is conducted on the vessel's navigational operation characteristics and their physical relationships using a large set of sensor data collected over a period of approximately two years. Preprocessing techniques are applied to prepare the data for modeling, and different machine learning approaches are used and compared to find the best performing model. The goal is to provide real-time navigational aid, and the models can also be useful for vessel design and maintenance, as well as for autonomous navigation \cite{gu2021autonomous}.}

\rv{In addition, we find that the LR family of models is suitable for non-dynamic systems with well-defined linear parameters, and their performance depends on the availability and quality of the features. Due to the complexity of the system dynamics and the deficiency in the features, all of these linear approaches suffer from poor validation loss. Non-parametric approaches can be useful for unknown model structures, and their performance can be enhanced by using an ensemble of techniques. However, their complexity and computational time are downsides. Neural network structures, such as MLP, for complex dynamic systems can provide relatively good performance, but they require careful normalization of input data \cite{agand2019adaptive}. We consider future predictions of vessel states for optimization purposes, with the goal of minimizing SFC under different weather and loading conditions. As a future direction, time-based models, such as LSTM, could be developed using instantaneous models and meta-heuristic approaches to provide navigational aid.}



\appendix
\section{Dataset}
\label{sec:appendixa}
\pa{Table \ref{tab:dataset} shows the definition variables in the dataset and their units. }
\begin{table}
\caption{List of dataset variables}
\begin{tabular*}{\tblwidth}{L@{}L@{}L@{}|L@{}L@{}L@{}}
\toprule
 Variable &Channel title &Unit & Variable & Channel title &Unit\\ 
\midrule
\textit{DEPTH} & depth of water & $m$ & \textit{ENGINE\_i\_FLOWRATE} & Engine $i=1,2$ flow rate & $l/min$ \\
\textit{ENGINE\_i\_RATE\_A} &Engine  $i=1,2$ flow rate A & $l/min$ &\textit{ENGINE\_i\_TEMP\_A} &Engine $i=1,2$ flow temp A & $^\circ C$\\
\textit{ENGINE\_i\_SFC} &Engine  $i=1,2$ SFC & $kg/h$ & \textit{HEADING}&Heading & $deg$\\
\textit{LATITUDE}&Latitude & $deg$ &\textit{LONGITUDE}& Longitude &$deg$\\
\textit{PITCH\_i} & Pitch $i=1,2$ & $\% $ &\textit{POWER\_i}& Power $i=1,2$ & $kW$\\
\textit{RATE\_OF\_TURN} & Rate of heading turn & $deg/s$ & \textit{SOG}& Speed over ground&$knots$\\
\textit{SOG\_LONG}&SOG longitudinal  speed& $knots$& \textit{SOG\_TRANS} & SOG transverse speed& $knots$\\
\textit{SPEED\_i} &Engine speed $i=1,2$  & $rpm$ & \textit{STW} & Speed through water &$knots$\\
\textit{THRUST\_i}&Engine thrust $i=1,2$ & $kN$&\textit{TORQUE\_i}&Engine torque $i=1,2$ &$kNm$\\
\textit{TRACK\_MADE\_GOOD}& Track made& $deg$ & \textit{WIND\_ANGLE} &Wind angle &$deg$\\
\textit{WIND\_SPEED}& Wind speed &$knots$ &\textit{WIND\_ANGLE\_TRUE} & Adjusted wind angle &$deg$ \\
\textit{WIND\_SPEED\_TRUE}&Adjusted wind speed & $knot$& \textit{OPERATIONAL\_MODE} &  Mode $1,0$ binary & NA\\
 \textit{TRIP\_DURATION}& Trip duration & $min$&\textit{CARGO}&Cargo&$kg$\\
\bottomrule
\end{tabular*}
\label{tab:dataset}
\end{table}


\bibliographystyle{cas-model2-names}

\bibliography{cas-refs.bib}

\begin{thebibliography}{51}
\expandafter\ifx\csname natexlab\endcsname\relax\def\natexlab#1{#1}\fi
\providecommand{\url}[1]{\texttt{#1}}
\providecommand{\href}[2]{#2}
\providecommand{\path}[1]{#1}
\providecommand{\DOIprefix}{doi:}
\providecommand{\ArXivprefix}{arXiv:}
\providecommand{\URLprefix}{URL: }
\providecommand{\Pubmedprefix}{pmid:}
\providecommand{\doi}[1]{\href{http://dx.doi.org/#1}{\path{#1}}}
\providecommand{\Pubmed}[1]{\href{pmid:#1}{\path{#1}}}
\providecommand{\bibinfo}[2]{#2}
\ifx\xfnm\relax \def\xfnm[#1]{\unskip,\space#1}\fi
\bibitem[{ABS(2013)}]{abs22}
\bibinfo{author}{ABS}, \bibinfo{year}{2013}.
\newblock \bibinfo{title}{Ship energy efficiency measures advisory}.
\newblock \bibinfo{howpublished}{Website}.
\newblock \URLprefix
  \url{https://ww2.eagle.org/content/dam/eagle/advisories-and-debriefs/ABS_Energy_Efficiency_Advisory.pdf}.
  \bibinfo{note}{(accessed on 7 July 2022)}.
\bibitem[{Agand et~al.(2023)Agand, Chen and Taghirad}]{agand2023online}
\bibinfo{author}{Agand, P.}, \bibinfo{author}{Chen, M.},
  \bibinfo{author}{Taghirad, H.D.}, \bibinfo{year}{2023}.
\newblock \bibinfo{title}{Online probabilistic model identification using
  adaptive recursive mcmc}, in: \bibinfo{booktitle}{2023 International Joint
  Conference on Neural Networks (IJCNN)}, \bibinfo{organization}{IEEE}. pp.
  \bibinfo{pages}{1--8}.
\bibitem[{Agand and Shoorehdeli(2019)}]{agand2019adaptive}
\bibinfo{author}{Agand, P.}, \bibinfo{author}{Shoorehdeli, M.A.},
  \bibinfo{year}{2019}.
\newblock \bibinfo{title}{Adaptive model learning of neural networks with uub
  stability for robot dynamic estimation}, in: \bibinfo{booktitle}{2019
  International Joint Conference on Neural Networks (IJCNN)},
  \bibinfo{organization}{IEEE}. pp. \bibinfo{pages}{1--6}.
\bibitem[{Agand et~al.(2016)Agand, Shoorehdeli and
  Teshnehlab}]{agand2016transparent}
\bibinfo{author}{Agand, P.}, \bibinfo{author}{Shoorehdeli, M.A.},
  \bibinfo{author}{Teshnehlab, M.}, \bibinfo{year}{2016}.
\newblock \bibinfo{title}{Transparent and flexible neural network structure for
  robot dynamics identification}, in: \bibinfo{booktitle}{2016 24th Iranian
  Conference on Electrical Engineering (ICEE)}, \bibinfo{organization}{IEEE}.
  pp. \bibinfo{pages}{1700--1705}.
\bibitem[{Agand et~al.(2022)Agand, Taherahmadi, Lim and Chen}]{agand2022human}
\bibinfo{author}{Agand, P.}, \bibinfo{author}{Taherahmadi, M.},
  \bibinfo{author}{Lim, A.}, \bibinfo{author}{Chen, M.}, \bibinfo{year}{2022}.
\newblock \bibinfo{title}{Human navigational intent inference with
  probabilistic and optimal approaches}, in: \bibinfo{booktitle}{2022
  International Conference on Robotics and Automation (ICRA)},
  \bibinfo{organization}{IEEE}. pp. \bibinfo{pages}{8562--8568}.
\bibitem[{Barnett and Lewis(1984)}]{barnett1984outliers}
\bibinfo{author}{Barnett, V.}, \bibinfo{author}{Lewis, T.},
  \bibinfo{year}{1984}.
\newblock \bibinfo{title}{Outliers in statistical data}.
\newblock \bibinfo{journal}{Wiley Series in Probability and Mathematical
  Statistics. Applied Probability and Statistics} .
\bibitem[{Be{\c{s}}ik{\c{c}}i et~al.(2016)Be{\c{s}}ik{\c{c}}i, Arslan, Turan
  and {\"O}l{\c{c}}er}]{becsikcci2016artificial}
\bibinfo{author}{Be{\c{s}}ik{\c{c}}i, E.B.}, \bibinfo{author}{Arslan, O.},
  \bibinfo{author}{Turan, O.}, \bibinfo{author}{{\"O}l{\c{c}}er, A.I.},
  \bibinfo{year}{2016}.
\newblock \bibinfo{title}{An artificial neural network based decision support
  system for energy efficient ship operations}.
\newblock \bibinfo{journal}{Computers \& Operations Research}
  \bibinfo{volume}{66}, \bibinfo{pages}{393--401}.
\bibitem[{Bialystocki and Konovessis(2016)}]{bialystocki2016estimation}
\bibinfo{author}{Bialystocki, N.}, \bibinfo{author}{Konovessis, D.},
  \bibinfo{year}{2016}.
\newblock \bibinfo{title}{On the estimation of ship's fuel consumption and
  speed curve: A statistical approach}.
\newblock \bibinfo{journal}{Journal of Ocean Engineering and Science}
  \bibinfo{volume}{1}, \bibinfo{pages}{157--166}.
\bibitem[{Bocchetti et~al.(2015)Bocchetti, Lepore, Palumbo and
  Vitiello}]{bocchetti2015statistical}
\bibinfo{author}{Bocchetti, D.}, \bibinfo{author}{Lepore, A.},
  \bibinfo{author}{Palumbo, B.}, \bibinfo{author}{Vitiello, L.},
  \bibinfo{year}{2015}.
\newblock \bibinfo{title}{A statistical approach to ship fuel consumption
  monitoring}.
\newblock \bibinfo{journal}{Journal of Ship Research} \bibinfo{volume}{59},
  \bibinfo{pages}{162--171}.
\bibitem[{Borkowski et~al.(2011)Borkowski, Kasyk and
  Kowalak}]{borkowski2011assessment}
\bibinfo{author}{Borkowski, T.}, \bibinfo{author}{Kasyk, L.},
  \bibinfo{author}{Kowalak, P.}, \bibinfo{year}{2011}.
\newblock \bibinfo{title}{Assessment of ship's engine effective power, fuel
  consumption and emission using the vessel speed}.
\newblock \bibinfo{journal}{Journal of KONES} \bibinfo{volume}{18},
  \bibinfo{pages}{31--39}.
\bibitem[{Branch and Stopford(2013)}]{branch2013maritime}
\bibinfo{author}{Branch, A.}, \bibinfo{author}{Stopford, M.},
  \bibinfo{year}{2013}.
\newblock \bibinfo{title}{Maritime economics}.
\newblock \bibinfo{publisher}{Routledge}.
\bibitem[{Dykstra and Heinrich(1996)}]{dykstra1996forest}
\bibinfo{author}{Dykstra, D.P.}, \bibinfo{author}{Heinrich, R.},
  \bibinfo{year}{1996}.
\newblock \bibinfo{title}{Forest Codes of Practice: Contributing to
  Environmentally Sound Forest Operations: Proceedings of an FAO/IUFRO Meeting
  of Experts on Forest Practices, Feldafing, Germany, 11-14 December 1994}.
\newblock \bibinfo{number}{133}, \bibinfo{publisher}{Food \& Agriculture Org.}
\bibitem[{Eide et~al.(2011)Eide, Longva, Hoffmann, Endresen and
  Dals{\o}ren}]{eide2011future}
\bibinfo{author}{Eide, M.S.}, \bibinfo{author}{Longva, T.},
  \bibinfo{author}{Hoffmann, P.}, \bibinfo{author}{Endresen, {\O}.},
  \bibinfo{author}{Dals{\o}ren, S.B.}, \bibinfo{year}{2011}.
\newblock \bibinfo{title}{Future cost scenarios for reduction of ship co2
  emissions}.
\newblock \bibinfo{journal}{Maritime Policy \& Management}
  \bibinfo{volume}{38}, \bibinfo{pages}{11--37}.
\bibitem[{Farag and {\"O}l{\c{c}}er(2020)}]{farag2020development}
\bibinfo{author}{Farag, Y.B.}, \bibinfo{author}{{\"O}l{\c{c}}er, A.I.},
  \bibinfo{year}{2020}.
\newblock \bibinfo{title}{The development of a ship performance model in
  varying operating conditions based on ann and regression techniques}.
\newblock \bibinfo{journal}{Ocean Engineering} \bibinfo{volume}{198},
  \bibinfo{pages}{106972}.
\bibitem[{Farkas et~al.(2020)Farkas, Song, Degiuli, Marti{\'c} and
  Demirel}]{farkas2020impact}
\bibinfo{author}{Farkas, A.}, \bibinfo{author}{Song, S.},
  \bibinfo{author}{Degiuli, N.}, \bibinfo{author}{Marti{\'c}, I.},
  \bibinfo{author}{Demirel, Y.K.}, \bibinfo{year}{2020}.
\newblock \bibinfo{title}{Impact of biofilm on the ship propulsion
  characteristics and the speed reduction}.
\newblock \bibinfo{journal}{Ocean Engineering} \bibinfo{volume}{199},
  \bibinfo{pages}{107033}.
\bibitem[{Gkerekos and Lazakis(2020)}]{gkerekos2020novel}
\bibinfo{author}{Gkerekos, C.}, \bibinfo{author}{Lazakis, I.},
  \bibinfo{year}{2020}.
\newblock \bibinfo{title}{A novel, data-driven heuristic framework for vessel
  weather routing}.
\newblock \bibinfo{journal}{Ocean Engineering} \bibinfo{volume}{197},
  \bibinfo{pages}{106887}.
\bibitem[{Gkerekos et~al.(2019)Gkerekos, Lazakis and
  Theotokatos}]{gkerekos2019machine}
\bibinfo{author}{Gkerekos, C.}, \bibinfo{author}{Lazakis, I.},
  \bibinfo{author}{Theotokatos, G.}, \bibinfo{year}{2019}.
\newblock \bibinfo{title}{Machine learning models for predicting ship main
  engine fuel oil consumption: A comparative study}.
\newblock \bibinfo{journal}{Ocean Engineering} \bibinfo{volume}{188},
  \bibinfo{pages}{106282}.
\bibitem[{G{\'o}rski et~al.(2013)G{\'o}rski, Abramowicz-Gerigk and
  Burciu}]{gorski2013influence}
\bibinfo{author}{G{\'o}rski, W.}, \bibinfo{author}{Abramowicz-Gerigk, T.},
  \bibinfo{author}{Burciu, Z.}, \bibinfo{year}{2013}.
\newblock \bibinfo{title}{The influence of ship operational parameters on fuel
  consumption}.
\newblock \bibinfo{journal}{Zeszyty Naukowe/Akademia Morska w Szczecinie} .
\bibitem[{Gu et~al.(2021)Gu, Goez, Guajardo and Wallace}]{gu2021autonomous}
\bibinfo{author}{Gu, Y.}, \bibinfo{author}{Goez, J.C.},
  \bibinfo{author}{Guajardo, M.}, \bibinfo{author}{Wallace, S.W.},
  \bibinfo{year}{2021}.
\newblock \bibinfo{title}{Autonomous vessels: state of the art and potential
  opportunities in logistics}.
\newblock \bibinfo{journal}{International Transactions in Operational Research}
  \bibinfo{volume}{28}, \bibinfo{pages}{1706--1739}.
\bibitem[{Harris and Kennedy(2021)}]{harris2021low}
\bibinfo{author}{Harris, T.}, \bibinfo{author}{Kennedy, A.},
  \bibinfo{year}{2021}.
\newblock \bibinfo{title}{Low friction recoating performance improvements
  aboard a passenger ferry}, in: \bibinfo{booktitle}{SNAME Maritime
  Convention}, \bibinfo{organization}{OnePetro}.
\bibitem[{Hastie et~al.(2001)Hastie, Tibshirani and
  Friedman}]{hastie2001elements}
\bibinfo{author}{Hastie, T.}, \bibinfo{author}{Tibshirani, R.},
  \bibinfo{author}{Friedman, J.}, \bibinfo{year}{2001}.
\newblock \bibinfo{title}{The elements of statistical learning. springer series
  in statistics}.
\newblock \bibinfo{journal}{New York, NY, USA} .
\bibitem[{Hastie et~al.(2009)Hastie, Tibshirani, Friedman and
  Friedman}]{hastie2009elements}
\bibinfo{author}{Hastie, T.}, \bibinfo{author}{Tibshirani, R.},
  \bibinfo{author}{Friedman, J.H.}, \bibinfo{author}{Friedman, J.H.},
  \bibinfo{year}{2009}.
\newblock \bibinfo{title}{The elements of statistical learning: data mining,
  inference, and prediction}. volume~\bibinfo{volume}{2}.
\newblock \bibinfo{publisher}{Springer}.
\bibitem[{Hu et~al.(2021)Hu, Zhou, Osman, Li, Jin and Zhen}]{hu2021novel}
\bibinfo{author}{Hu, Z.}, \bibinfo{author}{Zhou, T.}, \bibinfo{author}{Osman,
  M.T.}, \bibinfo{author}{Li, X.}, \bibinfo{author}{Jin, Y.},
  \bibinfo{author}{Zhen, R.}, \bibinfo{year}{2021}.
\newblock \bibinfo{title}{A novel hybrid fuel consumption prediction model for
  ocean-going container ships based on sensor data}.
\newblock \bibinfo{journal}{Journal of Marine Science and Engineering}
  \bibinfo{volume}{9}, \bibinfo{pages}{449}.
\bibitem[{Hu et~al.(2022)Hu, Zhou, Zhen, Jin, Li and Osman}]{hu2022two}
\bibinfo{author}{Hu, Z.}, \bibinfo{author}{Zhou, T.}, \bibinfo{author}{Zhen,
  R.}, \bibinfo{author}{Jin, Y.}, \bibinfo{author}{Li, X.},
  \bibinfo{author}{Osman, M.T.}, \bibinfo{year}{2022}.
\newblock \bibinfo{title}{A two-step strategy for fuel consumption prediction
  and optimization of ocean-going ships}.
\newblock \bibinfo{journal}{Ocean Engineering} \bibinfo{volume}{249},
  \bibinfo{pages}{110904}.
\bibitem[{Joung et~al.(2020)Joung, Kang, Lee and Ahn}]{joung2020imo}
\bibinfo{author}{Joung, T.H.}, \bibinfo{author}{Kang, S.G.},
  \bibinfo{author}{Lee, J.K.}, \bibinfo{author}{Ahn, J.}, \bibinfo{year}{2020}.
\newblock \bibinfo{title}{The imo initial strategy for reducing greenhouse gas
  (ghg) emissions, and its follow-up actions towards 2050}.
\newblock \bibinfo{journal}{Journal of International Maritime Safety,
  Environmental Affairs, and Shipping} \bibinfo{volume}{4},
  \bibinfo{pages}{1--7}.
\bibitem[{Karagiannidis and Themelis(2021)}]{karagiannidis2021data}
\bibinfo{author}{Karagiannidis, P.}, \bibinfo{author}{Themelis, N.},
  \bibinfo{year}{2021}.
\newblock \bibinfo{title}{Data-driven modelling of ship propulsion and the
  effect of data pre-processing on the prediction of ship fuel consumption and
  speed loss}.
\newblock \bibinfo{journal}{Ocean Engineering} \bibinfo{volume}{222},
  \bibinfo{pages}{108616}.
\bibitem[{Kim et~al.(2021)Kim, Jung and Park}]{kim2021development}
\bibinfo{author}{Kim, Y.R.}, \bibinfo{author}{Jung, M.}, \bibinfo{author}{Park,
  J.B.}, \bibinfo{year}{2021}.
\newblock \bibinfo{title}{Development of a fuel consumption prediction model
  based on machine learning using ship in-service data}.
\newblock \bibinfo{journal}{Journal of Marine Science and Engineering}
  \bibinfo{volume}{9}, \bibinfo{pages}{137}.
\bibitem[{Leifsson et~al.(2008)Leifsson, S{\ae}varsd{\'o}ttir, Sigur{\dh}sson
  and V{\'e}steinsson}]{leifsson2008grey}
\bibinfo{author}{Leifsson, L.{\TH}.}, \bibinfo{author}{S{\ae}varsd{\'o}ttir,
  H.}, \bibinfo{author}{Sigur{\dh}sson, S.{\TH}.},
  \bibinfo{author}{V{\'e}steinsson, A.}, \bibinfo{year}{2008}.
\newblock \bibinfo{title}{Grey-box modeling of an ocean vessel for operational
  optimization}.
\newblock \bibinfo{journal}{Simulation Modelling Practice and Theory}
  \bibinfo{volume}{16}, \bibinfo{pages}{923--932}.
\bibitem[{Meng et~al.(2016)Meng, Du and Wang}]{meng2016shipping}
\bibinfo{author}{Meng, Q.}, \bibinfo{author}{Du, Y.}, \bibinfo{author}{Wang,
  Y.}, \bibinfo{year}{2016}.
\newblock \bibinfo{title}{Shipping log data based container ship fuel
  efficiency modeling}.
\newblock \bibinfo{journal}{Transportation Research Part B: Methodological}
  \bibinfo{volume}{83}, \bibinfo{pages}{207--229}.
\bibitem[{Nelles(2020)}]{nelles2020nonlinear}
\bibinfo{author}{Nelles, O.}, \bibinfo{year}{2020}.
\newblock \bibinfo{title}{Nonlinear system identification: from classical
  approaches to neural networks, fuzzy models, and gaussian processes}.
\newblock \bibinfo{publisher}{Springer Nature}.
\bibitem[{Nielsen et~al.(2019)Nielsen, Sandvik, Pedersen, Asbj{\o}rnslett and
  Fagerholt}]{nielsen2019impact}
\bibinfo{author}{Nielsen, J.B.}, \bibinfo{author}{Sandvik, E.},
  \bibinfo{author}{Pedersen, E.}, \bibinfo{author}{Asbj{\o}rnslett, B.E.},
  \bibinfo{author}{Fagerholt, K.}, \bibinfo{year}{2019}.
\newblock \bibinfo{title}{Impact of simulation model fidelity and simulation
  method on ship operational performance evaluation in sea passage scenarios}.
\newblock \bibinfo{journal}{Ocean Engineering} \bibinfo{volume}{188},
  \bibinfo{pages}{106268}.
\bibitem[{Noufal and Hassan(2016)}]{noufal2016impact}
\bibinfo{author}{Noufal, C.A.S.H.}, \bibinfo{author}{Hassan, C.M.H.N.},
  \bibinfo{year}{2016}.
\newblock \bibinfo{title}{The impact of implementing the international
  convention on the control of harmful anti-fouling systems in ships (afs
  convention) on the marine environment}.
\newblock \bibinfo{journal}{Journal of Shipping and Ocean Engineering}
  \bibinfo{volume}{6}, \bibinfo{pages}{57--63}.
\bibitem[{Panapakidis et~al.(2020)Panapakidis, Sourtzi and
  Dagoumas}]{panapakidis2020forecasting}
\bibinfo{author}{Panapakidis, I.}, \bibinfo{author}{Sourtzi, V.M.},
  \bibinfo{author}{Dagoumas, A.}, \bibinfo{year}{2020}.
\newblock \bibinfo{title}{Forecasting the fuel consumption of passenger ships
  with a combination of shallow and deep learning}.
\newblock \bibinfo{journal}{Electronics} \bibinfo{volume}{9},
  \bibinfo{pages}{776}.
\bibitem[{Peng et~al.(2020)Peng, Liu, Li, Huang and Wang}]{peng2020machine}
\bibinfo{author}{Peng, Y.}, \bibinfo{author}{Liu, H.}, \bibinfo{author}{Li,
  X.}, \bibinfo{author}{Huang, J.}, \bibinfo{author}{Wang, W.},
  \bibinfo{year}{2020}.
\newblock \bibinfo{title}{Machine learning method for energy consumption
  prediction of ships in port considering green ports}.
\newblock \bibinfo{journal}{Journal of Cleaner Production}
  \bibinfo{volume}{264}, \bibinfo{pages}{121564}.
\bibitem[{Petersen(2011)}]{petersen2011mining}
\bibinfo{author}{Petersen, J.P.}, \bibinfo{year}{2011}.
\newblock \bibinfo{title}{Mining of ship operation data for energy
  conservation} .
\bibitem[{Rudzki and Tarelko(2016)}]{rudzki2016decision}
\bibinfo{author}{Rudzki, K.}, \bibinfo{author}{Tarelko, W.},
  \bibinfo{year}{2016}.
\newblock \bibinfo{title}{A decision-making system supporting selection of
  commanded outputs for a ship's propulsion system with a controllable pitch
  propeller}.
\newblock \bibinfo{journal}{Ocean Engineering} \bibinfo{volume}{126},
  \bibinfo{pages}{254--264}.
\bibitem[{Scott(2014)}]{bc20}
\bibinfo{author}{Scott}, \bibinfo{year}{2014}.
\newblock \bibinfo{title}{Bc ferries - queen of oak bay outbound from berth 2,
  departure bay.}
\newblock \bibinfo{howpublished}{Website}.
\newblock \URLprefix \url{https://www.flickr.com/photos/bcfs/12297338034}.
  \bibinfo{note}{(accessed on 28 August 2022)}.
\bibitem[{Soleymani et~al.(2018)Soleymani, Sharifi, Edalat, Sharifi and
  Zadeh}]{soleymani2018linear}
\bibinfo{author}{Soleymani, A.}, \bibinfo{author}{Sharifi, S.M.H.},
  \bibinfo{author}{Edalat, P.}, \bibinfo{author}{Sharifi, S.M.M.},
  \bibinfo{author}{Zadeh, S.K.}, \bibinfo{year}{2018}.
\newblock \bibinfo{title}{Linear modeling of marine vessels fuel consumption
  for ration of subsidized fuel}.
\newblock \bibinfo{journal}{INTERNATIONAL JOURNAL OF MARITIME
  TECHNOLOGY,[online]} \bibinfo{volume}{10}, \bibinfo{pages}{7--13}.
\bibitem[{Sudi~Mwasinago et~al.(2021)Sudi~Mwasinago, Siele and
  Agak}]{sudi2021effect}
\bibinfo{author}{Sudi~Mwasinago, A.}, \bibinfo{author}{Siele, R.},
  \bibinfo{author}{Agak, T.}, \bibinfo{year}{2021}.
\newblock \bibinfo{title}{Effect of foreign exchange rate on maritime sector
  performance in enhancing economic growth in kenya} .
\bibitem[{Tare{\l}ko(2014)}]{tarelko2014effect}
\bibinfo{author}{Tare{\l}ko, W.}, \bibinfo{year}{2014}.
\newblock \bibinfo{title}{The effect of hull biofouling on parameters
  characterising ship propulsion system efficiency}.
\newblock \bibinfo{journal}{Polish Maritime Research} \bibinfo{volume}{21},
  \bibinfo{pages}{27--34X}.
\bibitem[{Theodoridis(2009)}]{theodoridis2009chapter}
\bibinfo{author}{Theodoridis, S.}, \bibinfo{year}{2009}.
\newblock \bibinfo{title}{Chapter 15-clustering algorithms iv in: Theodoridis
  s, koutroumbas k, editors}.
\newblock \bibinfo{journal}{Pattern Recognition (Fourth Edition). Boston:
  Academic Press} , \bibinfo{pages}{765--862}.
\bibitem[{Thorndike~Robert(1996)}]{thorndike1996belong}
\bibinfo{author}{Thorndike~Robert, L.}, \bibinfo{year}{1996}.
\newblock \bibinfo{title}{Who belong in the family}.
\newblock \bibinfo{journal}{Psychometrika} \bibinfo{volume}{18},
  \bibinfo{pages}{267--276}.
\bibitem[{Trodden et~al.(2015)Trodden, Murphy, Pazouki and
  Sargeant}]{trodden2015fuel}
\bibinfo{author}{Trodden, D.}, \bibinfo{author}{Murphy, A.},
  \bibinfo{author}{Pazouki, K.}, \bibinfo{author}{Sargeant, J.},
  \bibinfo{year}{2015}.
\newblock \bibinfo{title}{Fuel usage data analysis for efficient shipping
  operations}.
\newblock \bibinfo{journal}{Ocean Engineering} \bibinfo{volume}{110},
  \bibinfo{pages}{75--84}.
\bibitem[{Tsitsilonis and Theotokatos(2018)}]{tsitsilonis2018novel}
\bibinfo{author}{Tsitsilonis, K.M.}, \bibinfo{author}{Theotokatos, G.},
  \bibinfo{year}{2018}.
\newblock \bibinfo{title}{A novel systematic methodology for ship propulsion
  engines energy management}.
\newblock \bibinfo{journal}{Journal of cleaner production}
  \bibinfo{volume}{204}, \bibinfo{pages}{212--236}.
\bibitem[{Uyan{\i}k et~al.(2020)Uyan{\i}k, Karatu{\u{g}} and
  Arslano{\u{g}}lu}]{uyanik2020machine}
\bibinfo{author}{Uyan{\i}k, T.}, \bibinfo{author}{Karatu{\u{g}}, {\c{C}}.},
  \bibinfo{author}{Arslano{\u{g}}lu, Y.}, \bibinfo{year}{2020}.
\newblock \bibinfo{title}{Machine learning approach to ship fuel consumption: A
  case of container vessel}.
\newblock \bibinfo{journal}{Transportation Research Part D: Transport and
  Environment} \bibinfo{volume}{84}, \bibinfo{pages}{102389}.
\bibitem[{Wang et~al.(2018)Wang, Ji, Zhao, Liu and Xu}]{wang2018predicting}
\bibinfo{author}{Wang, S.}, \bibinfo{author}{Ji, B.}, \bibinfo{author}{Zhao,
  J.}, \bibinfo{author}{Liu, W.}, \bibinfo{author}{Xu, T.},
  \bibinfo{year}{2018}.
\newblock \bibinfo{title}{Predicting ship fuel consumption based on lasso
  regression}.
\newblock \bibinfo{journal}{Transportation Research Part D: Transport and
  Environment} \bibinfo{volume}{65}, \bibinfo{pages}{817--824}.
\bibitem[{Wong and Yeh(2019)}]{wong2019reliable}
\bibinfo{author}{Wong, T.T.}, \bibinfo{author}{Yeh, P.Y.},
  \bibinfo{year}{2019}.
\newblock \bibinfo{title}{Reliable accuracy estimates from k-fold cross
  validation}.
\newblock \bibinfo{journal}{IEEE Transactions on Knowledge and Data
  Engineering} \bibinfo{volume}{32}, \bibinfo{pages}{1586--1594}.
\bibitem[{Yan et~al.(2020)Yan, Wang and Du}]{yan2020development}
\bibinfo{author}{Yan, R.}, \bibinfo{author}{Wang, S.}, \bibinfo{author}{Du,
  Y.}, \bibinfo{year}{2020}.
\newblock \bibinfo{title}{Development of a two-stage ship fuel consumption
  prediction and reduction model for a dry bulk ship}.
\newblock \bibinfo{journal}{Transportation Research Part E: Logistics and
  Transportation Review} \bibinfo{volume}{138}, \bibinfo{pages}{101930}.
\bibitem[{Zheng et~al.(2019)Zheng, Zhang, Yin, Liang, Wang, Li, Song and
  Zhang}]{zheng2019voyage}
\bibinfo{author}{Zheng, J.}, \bibinfo{author}{Zhang, H.}, \bibinfo{author}{Yin,
  L.}, \bibinfo{author}{Liang, Y.}, \bibinfo{author}{Wang, B.},
  \bibinfo{author}{Li, Z.}, \bibinfo{author}{Song, X.}, \bibinfo{author}{Zhang,
  Y.}, \bibinfo{year}{2019}.
\newblock \bibinfo{title}{A voyage with minimal fuel consumption for cruise
  ships}.
\newblock \bibinfo{journal}{Journal of Cleaner Production}
  \bibinfo{volume}{215}, \bibinfo{pages}{144--153}.
\bibitem[{Zhou et~al.(2022)Zhou, Hu, Hu and Zhen}]{zhou2022adaptive}
\bibinfo{author}{Zhou, T.}, \bibinfo{author}{Hu, Q.}, \bibinfo{author}{Hu, Z.},
  \bibinfo{author}{Zhen, R.}, \bibinfo{year}{2022}.
\newblock \bibinfo{title}{An adaptive hyper parameter tuning model for ship
  fuel consumption prediction under complex maritime environments}.
\newblock \bibinfo{journal}{Journal of Ocean Engineering and Science}
  \bibinfo{volume}{7}, \bibinfo{pages}{255--263}.
\bibitem[{Zhu et~al.(2021)Zhu, Zuo and Li}]{zhu2021modeling}
\bibinfo{author}{Zhu, Y.}, \bibinfo{author}{Zuo, Y.}, \bibinfo{author}{Li, T.},
  \bibinfo{year}{2021}.
\newblock \bibinfo{title}{Modeling of ship fuel consumption based on
  multisource and heterogeneous data: Case study of passenger ship}.
\newblock \bibinfo{journal}{Journal of Marine Science and Engineering}
  \bibinfo{volume}{9}, \bibinfo{pages}{273}.

\end{thebibliography}



\end{document}